\documentclass{article}

\usepackage{arxiv}

\usepackage[utf8]{inputenc} 
\usepackage[T1]{fontenc}    
\usepackage{hyperref}       
\usepackage{url}            
\usepackage{booktabs}       
\usepackage{amsfonts}       
\usepackage{nicefrac}       
\usepackage{microtype}      
\usepackage{lipsum}
\usepackage{amsmath}
\usepackage{epstopdf} 
\usepackage{graphicx} 
\usepackage{float}
\usepackage{subcaption}
\usepackage{booktabs}
\usepackage{float}
\usepackage{multirow}
\usepackage{comment}
\usepackage{booktabs,caption}
\usepackage[flushleft]{threeparttable}
\usepackage{graphicx}
\usepackage{setspace}
\usepackage{natbib}
\setcitestyle{square,comma,numbers}
\usepackage{hyperref}
\usepackage{lipsum}

\floatstyle{plaintop}
\restylefloat{table}

\title{An Automated Spectral Clustering for Multi-scale Data}
\date{\vspace{-5ex}}

\author{
 Milad Afzalan\\
 Graduate Research Assistant\\
  Department of Civil and Environemntal Engineering\\
  Virginia Tech\\
  Blacksburg, VA, 24060 \\
  \texttt{afzalan@vt.edu} \\
   \And
 Farrokh Jazizadeh \\
 Assistant Professor\\
  Department of Civil and Environemntal Engineering\\
  Virginia Tech\\
  Blacksburg, VA, 24060 \\
  \texttt{jazizade@vt.edu} \\
}

\begin{document}
\maketitle

\begin{abstract}
Spectral clustering algorithms typically require \emph{a priori} selection of input parameters such as the number of clusters, a scaling parameter for the affinity measure, or ranges of these values for parameter tuning. Despite efforts for automating the process of spectral clustering, the task of grouping data in multi-scale and higher dimensional spaces is yet to be explored. This study presents a spectral clustering heuristic algorithm that obviates the need for an input by estimating the parameters from the data itself. Specifically, it introduces the heuristic of iterative eigengap search with (1) global scaling and (2) local scaling. These approaches estimate the scaling parameter and implement iterative eigengap quantification along a search tree to reveal dissimilarities at different scales of a feature space and identify clusters. The performance of these approaches has been tested on various real-world datasets of power variation with multi-scale nature and gene expression. Our findings show that iterative eigengap search with a PCA-based global scaling scheme can discover different patterns with an accuracy of higher than 90\% in most cases without asking for a priori input information.
\end{abstract}


\section{Introduction}
Clustering, the practice of partitioning data into different groups with similar observations, has a variety of applications in knowledge discovery for unknown phenomena in different fields such as object recognition \cite{1,2}, cyber-physical systems \cite{3,4}, or bioinformatics \cite{5}. Spectral clustering \cite{6,7,8} is a data analytics technique that has gained popularity in recent years. Due to its capability of high-quality clustering and handling non-convex clusters that are typically challenging for other methods \cite{8}, spectral clustering has been implemented in different domains like computer vision and speech separation with promising performance \cite{9,10,11,12}. An overview of the literature reveals that spectral clustering has been \emph{adopted and adapted} for different application domains. In this study, we have explored automated spectral clustering for feature spaces with multi-scale and higher dimensional attributes. Our vision has stemmed from the need for self-configuring algorithms in cyber-physical systems that need to adapt their behavior in different settings. 

Spectral clustering decomposes the eigenvectors of a Laplacian matrix derived from an affinity matrix (i.e., similarity matrix) of the data and transforms the data into a new dimension, where it can be grouped with k-means or other algorithms that minimize a distortion metric. The affinity matrix in this context demonstrates the pairwise similarity between data points and is used to overcome the difficulties due to the lack of convexity in the data distribution. While considered as an unsupervised method, the algorithm calls for the determination of the number of clusters and a scaling parameter (that defines the behavior of the affinity matrix), which require algorithm tuning and \emph{a priori} data provision. These values are commonly provided based on data-driven parameter tuning (i.e., model selection) techniques or the general knowledge of a domain and therefore, make the autonomous application of algorithms more challenging. Accordingly, research efforts have been made on automated (also known as self-tuning) spectral clustering with a focus on particular problems. 

The main body of work in automated spectral clustering has focused on challenging two-dimensional and image segmentation problems (e.g., \cite{13,14,15,16}). Automated spectral clustering for multi-scale high dimensional data (mainly time-series) is another domain of research that has been less explored. Clustering in higher dimensions could be a challenging task \cite{17}; however, it has interesting applications in domains such as energy or power consumption pattern analysis, gene expression groupings, or speech separation. In other words, with emerging (and fast-growing) technologies for autonomous systems and smart environments, mainly in the form of cyber-physical systems, the need for self-tuning and context-aware algorithms that do not require human intervention for cluster analysis is increasing. The challenges for clustering of this type of data include: (1) Different groups of data can reside on different scales creating a multi-scale nature for the feature space. Consequently, larger scale components can mask the distinction of complex patterns in the smaller scales, and (2) the presence of noise in the acquired data could add to the complexity of the clustering process. Accordingly, in this study, we have proposed a heuristic algorithm for automated spectral clustering of multi-scale higher-dimensional data in order to obviate the need for \emph{a priori} information (i.e., number of cluster $K$, and scaling parameter $\sigma$) so that the algorithms could configure their behavior by learning from the data. 

Our proposed approach is built on the eigengap metric by introducing a new heuristic algorithm that couples eigengap with data-driven estimation of scaling parameter and a search framework that accounts for the multi-scale nature of the feature space. The performance of the proposed heuristic has been evaluated on real-world labeled datasets with multi-scale nature in a higher-dimensional space and compared to the performance of commonly used internal validation techniques that call for a threshold as the stopping criterion (i.e. the number of cluster optimization). The proposed method was initially motivated by the task of energy disaggregation, which is the practice of dividing the aggregate power series into individual appliance components with considerable power draw values (i.e., different scales). While there are recent well-known clustering studies for electricity energy monitoring (e.g. \cite{18,19}), they presume the number of appliances and only handle appliances with high power draws. However, due to our interest in the automation of cyber-physical systems, we are seeking to perform clustering without parameter-tuning or a priori information (i.e., number of clusters) provision. In addition to the application of the energy disaggregation, the approach has shown the potential to be applied to other similar data types as we have presented a benchmark gene expression clustering. 

The rest of the paper has been structured as follows: In section \ref{sec2}, a background on automated (i.e., self-tuning) spectral clustering as well as clustering of high dimensional data is presented. Section \ref{sec3} presents the proposed heuristic by introducing methods for estimation of scaling parameter and the number of clusters ($K$) that formalize our proposed framework for automated clustering. Section \ref{sec4} discusses the datasets and their properties and then proceeds with presenting the results, evaluation, and efficacy of the heuristic algorithms. Finally, the conclusion summarizes the work and its findings.

\section{Related Works in Spectral Clustering}
\label{sec2}
Spectral clustering has gained popularity due to their ease of implementation and efficiency in clustering \cite{20,21}. Therefore, in recent decades, several clustering algorithms have been proposed and used for different applications. The focus in these algorithms has been on the application of the similarity matrix spectrum for dimensionality reduction and feature space transformation to introduce convexity. One of the well-known algorithms in this field is the one proposed by Ng, Jordan, and Weiss (referred to as NJW) \cite{8}. In addition to the efforts in the formalization of spectral clustering algorithms, a number of studies have focused on expanding the algorithms into instances, which are capable of self-tuning or automated identification of natural partitions (or groups) in the data. Natural in this context refers to the clusters (or groups) that represent the actual/physical separation in the data. 

\subsection{Automated Spectral Clustering}
As a widely adopted technique, Zelnik-Manor and Perona introduced a self-tuning spectral clustering algorithm \cite{13} (built on NJW) that accounts for multiple scales in the feature space and automatically identifies the number of clusters using an optimization technique over a range of possible numbers for clusters. As part of this algorithm, they have proposed a novel similarity measure that integrates a data-driven scaling parameter by considering the distance of each point with some of its nearest neighbors. Scaling parameter refers to a parameter that controls the width of the neighborhood in the similarity metric. The number of clusters was estimated through examining a range of possible group numbers, recovering the rotation that best aligns the eigenvector of the matrix obtained from the data, and minimizing a cost function for possible rotations. The algorithm’s performance has been evaluated on a number of reference 2D problems (identified as benchmarks) as well as image segmentation problems, with promising performance.

A major part of the efforts in the field of automated spectral clustering has focused on the problem of image segmentation and thus the aforementioned study (i.e., \cite{13}) has been used as the benchmark for comparative analyses. In a class of these studies, it has been argued that the eigenvector selection is a crucial task for clustering because not all of the largest vectors are informative for natural segmentation of the data. These studies mainly sought the task of automatic determination of the number of clusters under noisy and sparse data. Different methods have been proposed to account for eigenvector selection. Identifying the relevance of the eigenvectors according to their contribution in determining the number of clusters \cite{14} and eigenvector selection through direct entropy ranking or a combination of elements in the ranking \cite{15} are examples of these methods. Other studies have proposed alternative solutions to address the problem of automated clustering in image segmentation. For example, \cite{16} uses non-normalized information of eigenvectors (rather than using a unit space for feature representation) and \cite{22} performed iterative cluster and merge in order to address the problem of image down-sizing (which can lead to losing fine details). 

Unlike the aforementioned efforts that have proposed solutions for challenging 2D datasets and image segmentation, \cite{23} proposed a kernel spectral clustering for a large-scale network without parameter input. To this end, entropy was used to detect the block-diagonal of the affinity matrix that was created by the projections in the eigenspace. The efficacy of the proposed approach was studied through synthetic data and real-world network datasets. While these existing approaches \cite{14,16,22,23} were developed to tackle spectral clustering in an automated manner, they are either designed to solve the problem for multi-scale 2D and image segmentation or network data, which is different in nature from data with multi-scale higher dimensional attributes as sought here.

\subsection{Spectral Clustering in Higher Dimension}
Spectral clustering for higher dimensional feature spaces has also been the subject of some studies (e.g. \cite{24,25,26,27,28,29}) to address different challenges. One of the examples of higher dimensional spaces in the real-world application is the time-series data. High level of noise and uneven sequence of length in data representation were among the challenges that have been taken into account. A class of studies has coupled spectral clustering and hidden Markov models (HMM) to benefit from structure and parametric assumptions of HMMs. These algorithms were evaluated on real-world datasets of motion capture, handwriting time-series sequence, sign language, and noisy sensor network data (e.g., \cite{24}, \cite{30}). The inevitable challenge of noise in real-world data has led to studies on spectral clustering approaches that are robust to noise. Examples of techniques that focused on robustness to noise include using a mapping approach based on regularization into a new space to separate the noise points in a new cluster \cite{25}, and proposing a partitioning criterion (discriminative hypergraph) which considers the intra-cluster compactness and inter-cluster separation of vertices \cite{31}. The performance of these studies was evaluated on datasets including digit numbers with 256 features and gene expression data. In another class of studies with higher dimensional features, clustering of large-scale datasets (both in the number of features and instances) were explored since they are computationally expensive \cite{26,27,28,29,32}. These approaches typically integrate sparse coding-based graph or apply approximation methods to reduce cost while the performance might be deteriorated. Among the works that attempted to enhance the performance, we can mention the application of a landmark-based spectral clustering \cite{28} that selects representative data points so that original data points are the linear combination of these landmarks and utilization of a sparse matrix and local interpolation to improve the approximate outputs \cite{29}. These studies had a focus on the efficiency and improved performance of the algorithm or been applied toward a specific application solution. Therefore, they have considered parameter selection and prior knowledge of the domain.

Considering the existence of real-world data with higher dimensional attributes, our study focuses on a heuristic spectral clustering algorithm that can robustly reveal different groupings for a class of multi-scale data for autonomous systems that need to adapt to different contexts. 

\section{An Automated Spectral Clustering Heuristic} \label{sec3}
The fundamentals of spectral clustering methods have been extensively described in the literature (e.g., \cite{8}, \cite{20}, \cite{13}, \cite{33,34}). Our heuristics is built on the NJW spectral clustering algorithm \cite{8}. A brief description of the NJW algorithm is followed to expand on it for our extended algorithm.

Assuming the data set $S=[s_1,s_2,s_3…,s_n ] \in R^{ n \times m}$ with $K$ clusters, the NJW algorithm steps are as follows:

1) Develop the affinity matrix $A\in R^{ n \times m}$, defined by:

\begin{equation}
A_{ij} = \begin{cases}
                 exp(-\frac{	\parallel s_i-s_j 	\parallel}{2\sigma^2}) & i \neq j\\
                 0 & i=j\\
              \end{cases}
\label{equation1}
\end{equation}
where $\sigma^2$ is the scaling parameter of the model.

2) Using $D$, a diagonal matrix with the summation of the elements on the $i$-th ($i \in [1,2,…,n]$) row of $A$ as $D(i,i)$, the Laplacian matrix is defined as:

\begin{equation}
L =  D^\frac{-1}{2} A D ^ \frac{1}{2}
\label{equation2}
\end{equation}

3) Compute $v_1,v_2,…,v_K$ the $K$ largest eigenvectors of $L$, and form the matrix $V=[v_1,v_2,…,v_K ] \in R^{n \times K}$

4) Form matrix $Y \in R^{n \times K}$ by renormalizing each row of $V$ as:

\begin{equation}
Y_{ij} = \frac{V_{ij}}{(\sum_{j} V_{ij}^2)^\frac{1}{2}}
\label{equation3}
\end{equation}

5) Cluster each row of $Y$ as a point in $R^K$ via K-means algorithm.

6) Original point $s_i$ belongs to cluster $k$, if and only if row $i$ of the matrix $Y$ is assigned to $k$. 

As the above steps state, the algorithm calls for input information. This information includes (1) the number of clusters ($K$), similar to other clustering algorithms that either need $K$ (e.g., the well-known K-means) or other input parameters such as thresholds as stopping criteria or model parameters (e.g., hierarchical clustering or mean-shift \cite{35,36}), and (2) a scaling parameter ($\sigma^2$) for forming the affinity matrix. These parameters could be estimated by human knowledge for specific problem domains or through internal validation, which also requires a threshold identification as an input parameter. Specifically, as proposed by Ng et al. \cite{8}, scaling parameter can be automatically fine-tuned by running the algorithm several times and selecting an optimal value from a range so that least distorted clusters of the rows in $Y$ are obtained. However, identifying this range calls for knowledge of the data, which contradicts the self-configuration objective.

\subsection{Estimating Scaling Parameter}
\begin{figure}
       \includegraphics[width=0.40\textwidth]{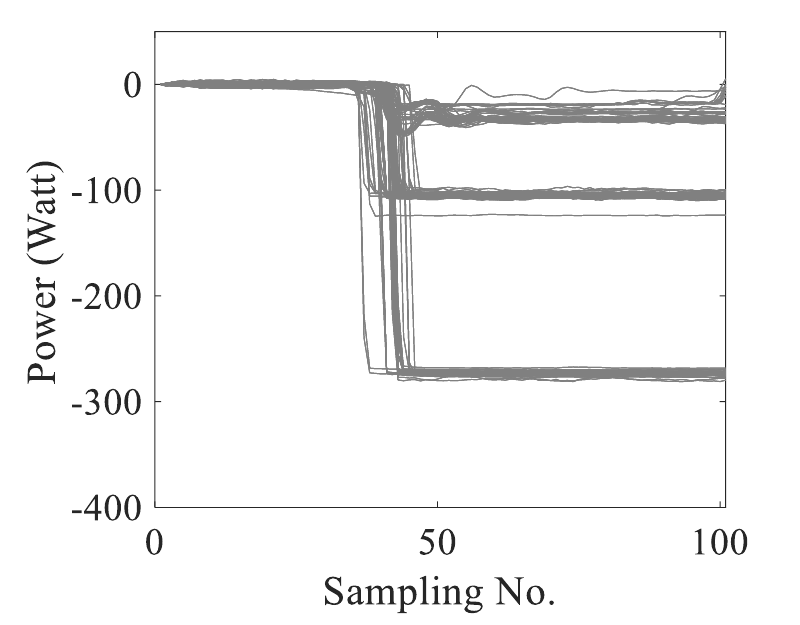}
     \hfill
       \includegraphics[width=0.40\textwidth]{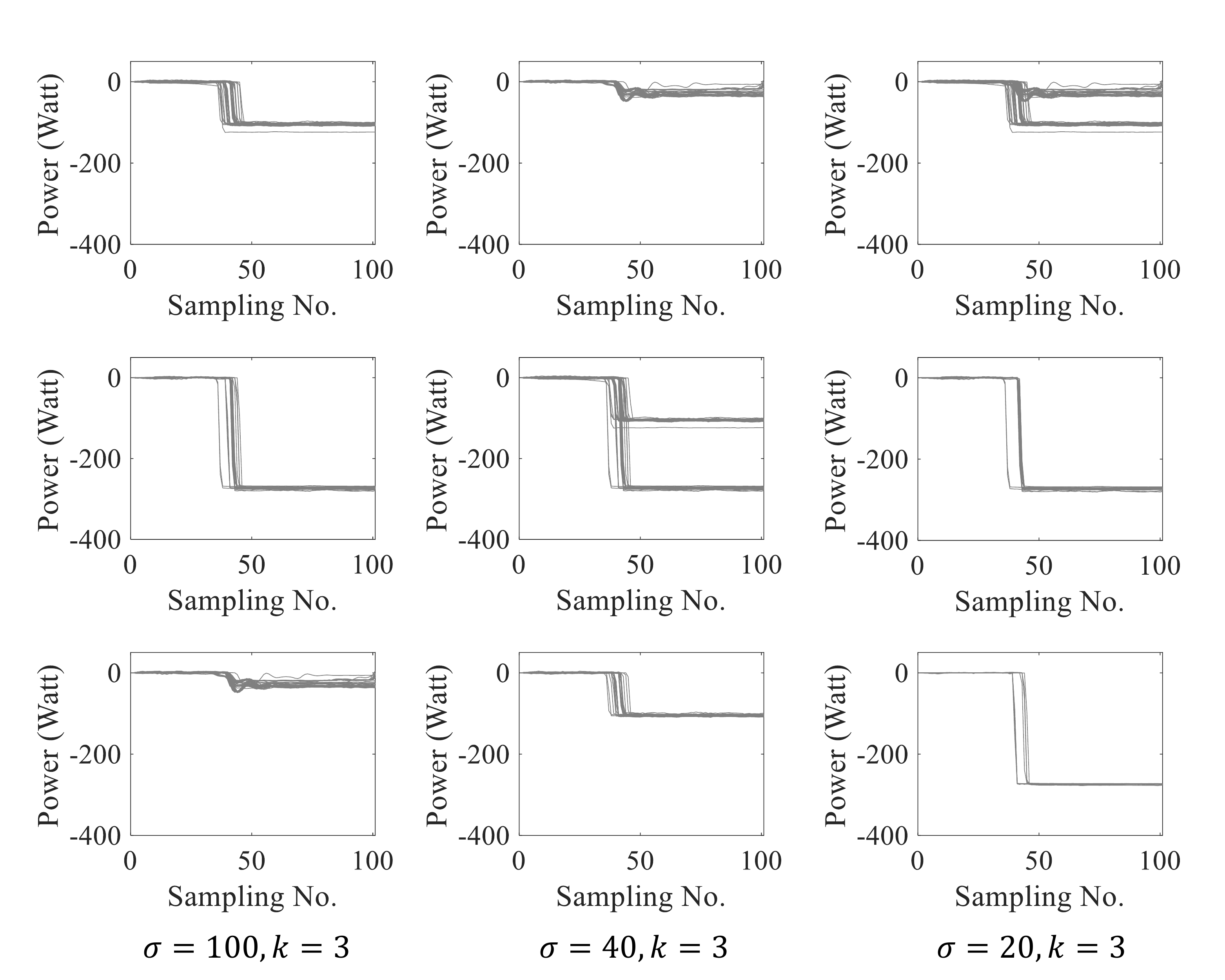}
     \caption{Sensitivity of $\sigma$ on clustering output for a power time-series data. (a) is the feature vectors for three appliances and (b) shows the clustering results for three different values of $\sigma$. ($\sigma$=100 gives the right prediction)}
     \label{fig1}
   \end{figure}

The scaling parameter ($\sigma^2$), shown in Eq. (\ref{equation1}), defines the width of neighborhoods which subsequently affects the calculation of the affinity matrix. In other words, it is a reference distance, below which two points are evaluated as similar and beyond which dissimilar \cite{13}, \cite{37}. Ng et al. \cite{8} describe the scaling parameter as the parameter that controls how rapidly the affinity falls off with the distance between two observations (i.e., data points). Therefore, selection of this parameter characterizes the dissimilarity in the feature space and thus the structure of clusters. In order to illustrate the effect of scaling parameter on the clustering, we have presented the outcome of spectral clustering on a power dataset, with high dimensional data points (which are subsections of a power time-series) in Fig \ref{fig1}. The data in this figure represent selected feature vectors for a problem of energy disaggregation, which uses signal processing and machine learning algorithms to identify the contribution of individual appliances on the aggregated power time-series. The time-series data is collected through one sensor on the main circuit panel in a building to avoid extensive instrumentation. Thus, this figure also shows the challenges of clustering in energy disaggregation. The spectral clustering outcome was visualized in Fig \ref{fig1}(b) for different $\sigma$ values to demonstrate the sensitivity. NJW algorithm was used with three as the number of clusters. While in all the cases the number of clusters is set to the correct number ($k$=3) as depicted in Fig \ref{fig1}(a), variation of $\sigma$ can affect the performance. As in this case, $\sigma$=100 is a suitable estimation while $\sigma$=20 or $\sigma=40$ leads to false prediction. The performance is sensitive to the scaling value over different datasets, indicating the importance of correct inference for the automated approach. Therefore, in our proposed heuristic, we have adopted the following methods for data-driven scaling factor estimation.

\subsubsection{PCA-based Scaling Parameter} \label{sec311}
Scaling parameter identifies the boundaries of the similarity neighborhood. A larger $\sigma$ indicates the similarity with more distant data points, whereas smaller values highlight the neighboring points. Therefore, in order to estimate the scale of neighborhoods, we have adopted the application of principal component analysis (PCA), which utilizes an orthogonal transformation to map the original variables into new space with uncorrelated variables. Given the higher dimensionality of data, we employ PCA in our approach to ensure that we focus on components of the feature space that account for the most variance in the data. In this study, through observations, we consider the one-time standard deviation of major principal axes (that accounts for the maximum variance in the whole data) in order to estimate the $\sigma$. This assumption allows us to form an approximate boundary threshold for distinguishing similar and dissimilar points based on the distribution of data points. Since only the first few components constitute the most variance, the number of considered principal axes is selected such that at least 95 percent of total variance is granted. This ensures reducing the input while also accounting for the whole variability of data.

Let us consider a set of data points $S$ with $n$ observations and $m$ features as:

\begin{equation}
S = [s_1,s_2,s_3,...,s_n],  \quad S \in R^{n \times m}
\label{equation4}
\end{equation}

The eigenvectors that correspond to the highest eigenvalues of the covariance of $S$ are associated with the highest variance. The projection matrix $U$ is formed by stacking the eigenvectors of corresponding eigenvalues sorted in the descending order. Using $U$, the sample data is transformed into the new space as follows:

\begin{equation}
P = S \times U,  \quad P \in R^{n \times m}
\label{equation5}
\end{equation}

Each principal component is derived by selecting the corresponding column from $P$. We employ the variance information from principal components for evaluating the scaling parameter. Therefore, the scaling parameter ($\sigma^2$) will be estimated as follows:

\begin{equation}
\sigma^2 = \frac{\sum_{i=1}^{y} w_i v_i}{\sum_{i=1}^{y} w_i }
\label{equation6}
\end{equation}

where $w_i$ is the data-driven weighting factor from $\pmb{w}$. $w_i$ denotes the ratio of the variance for the $i$-th principal axis to the summation of variance from all the principal axes (contained in $P$):

\begin{equation}
\pmb{w}^T=[w_1,w_2,...,w_m], \quad w_1>w_2>...>w_m
\label{equation7}
\end{equation}

We select $y$ major principal axes (in Eq. (\ref{equation6})) such that:

\begin{equation}
\frac{\sum_{i=1}^{y} w_i v_i}{\sum_{i=1}^{y} w_i }>0.95
\label{equation8}
\end{equation}

Here, $y$ is the smallest integer that satisfies the above inequality. The above inequality is used to consider almost the whole variability of data from PCA without considering all the principal axes. Typically, the first few principal axes account for the most variance in the data. An empirical analysis for Eq. (\ref{equation8}) is provided in section \ref{321}. \newline
Also, $\pmb{v}$ contains the variance of data points that are projected along the principal axes (each axis contains $n$ points, i.e., the total number of observations in the data).
\begin{equation}
\pmb{v}^T=[v_1,v_2,...,v_m], \quad v_1>v_2>...>v_m
\label{equation9}
\end{equation}

\subsubsection{Local scaling for self-tuning}
As a data-driven approach for estimation of the scaling parameter, Zelnik-Manor and Perona \cite{13} suggested that, instead of considering a global parameter for the whole affinity matrix, a local scale for each point allows the point-to-point distance self-tuning, which can further be used to compute the affinity for pairwise points. As proposed by \cite{13}, instead of using Eq. (\ref{equation1}) for calculating the affinity matrix, a local scale parameter is defined by each point, and Eq. (\ref{equation1}) is re-written as:

\begin{equation}
A_{ij} = exp(-\frac{	\parallel s_i-s_j 	\parallel}{\sigma_i \sigma_j})
\label{equation10}
\end{equation}

where 

\begin{equation}
\sigma_i = \parallel x_i-x_k 	\parallel
\label{equation11}
\end{equation}

$x_k$ is the $k$-th nearest neighbor of $x_i$. Through empirical observations, a value of $k$=7 was suggested \cite{13} that works for a range of applications. Local scaling provides a highly representative measure of scale for each data point. However, comparing to methods that use a global scale, this improvement in estimation comes with a higher computational cost since it calls for a KNN search for each data point in the process of forming the affinity matrix.

\subsection{Iterative Eigengap Search Heuristic}
Determining the number of clusters is a challenging problem even for human users and selecting the “right number of groups” is subject to different interpretation. In the case of spectral clustering, a commonly used heuristic is the eigengap that measures the stability of the eigenvectors in the Laplacian matrix. Based on the matrix perturbation theory, the subspace spanned by the first $i$ eigenvectors of the Laplacian matrix $L$ is stable if and only if the eigengap measure in Eq. (\ref{equation12}) is large:

\begin{equation}
\delta_i = 	\mid \lambda_i - \lambda_\text{i+1}	\mid
\label{equation12}
\end{equation}

where $N$ is the total number of observations, $\delta_1,…, \delta_\text{n-1}$ are the eigengaps, and $\lambda_1,...\lambda_n$ are the eigenvalues of the Laplacian matrix $L$ (defined in Eq. (\ref{equation2})). \newline
The value of eigengap for subsequent eigenvalues can indicate the place to pick the number of clusters. By examining the eigenvalue measures, the number of clusters can be estimated through:

\begin{equation}
K=argmax_i(\delta_i)
\label{equation13}
\end{equation}

where $\delta_i$ is calculated through Eq. (\ref{equation12}). 

Eigengap heuristic is a technique that mainly works in well-separated feature spaces \cite{10} but is not capable of properly partitioning the feature space in the presence of multi-scale data or mixed background as is the usual case in real-world data. In order to extend the capabilities of eigengap metric for multiscale data analysis, we propose to use an iterative eigengap search to reveal the complex topology of the feature space.

\subsubsection{Iterative Eigengap Search with global scale} \label{321}
In a multi-scale feature space, the data in the larger scale could mask the dissimilarities and therefore the microstructure of the smaller scales. However, if an algorithm explores the structure of feature space in different scales, the dissimilarities could be revealed and therefore, the eigengap metric could be utilized for identifying the number of clusters. This is the rationale behind our proposed Iterative Eigengap Search (IES) that partitions a feature space through searching along a tree-like structure. While at each iteration the eigengap might not find the final groupings of data points, it will segregate the data in different scales and consequently accentuates the dissimilarities at each scale. Therefore, the algorithm could refine the clusters through visiting each node (i.e., cluster) from the previous iteration by passing it to the spectral clustering algorithm. The process of refining the clusters will be carried out until eigengap cannot reveal finer structures in a leaf node. In other words, the stopping criterion is when all the nodes of the tree have been visited and all the leaf nodes contain only one cluster according to the eigengap measure.
Fig \ref{fig2} illustrates the conceptual process of Iterative Eigengap Search tree. In this tree structure, the root indicates the whole dataset, which is passed through eigengap heuristic (Eq.(\ref{equation12})) to determine an initial $K$ (Eq. (\ref{equation13})), and then clustered with NJW algorithm with a PCA-based global scaling parameter. Through empirical observations, we found that the $K$ should be sought in the first half of the vector of eigenvalues. Each of the produced nodes is processed with eigengap heuristic to be clustered again. As schematically shown in Fig \ref{fig2}, there are 6 nodes with estimated $k$=1 (after being analyzed with eigengap) that are accepted as final clusters, which all together form the data in the root node. The final clusters are thus the ones at the leaf nodes. At each level of the tree and for each node, the $\sigma$ measure is updated with respect to the content of that node to identify a scaling factor for that specific subset of data points. 

 \begin{figure}[H]
\centering
\includegraphics[width=0.7\textwidth]{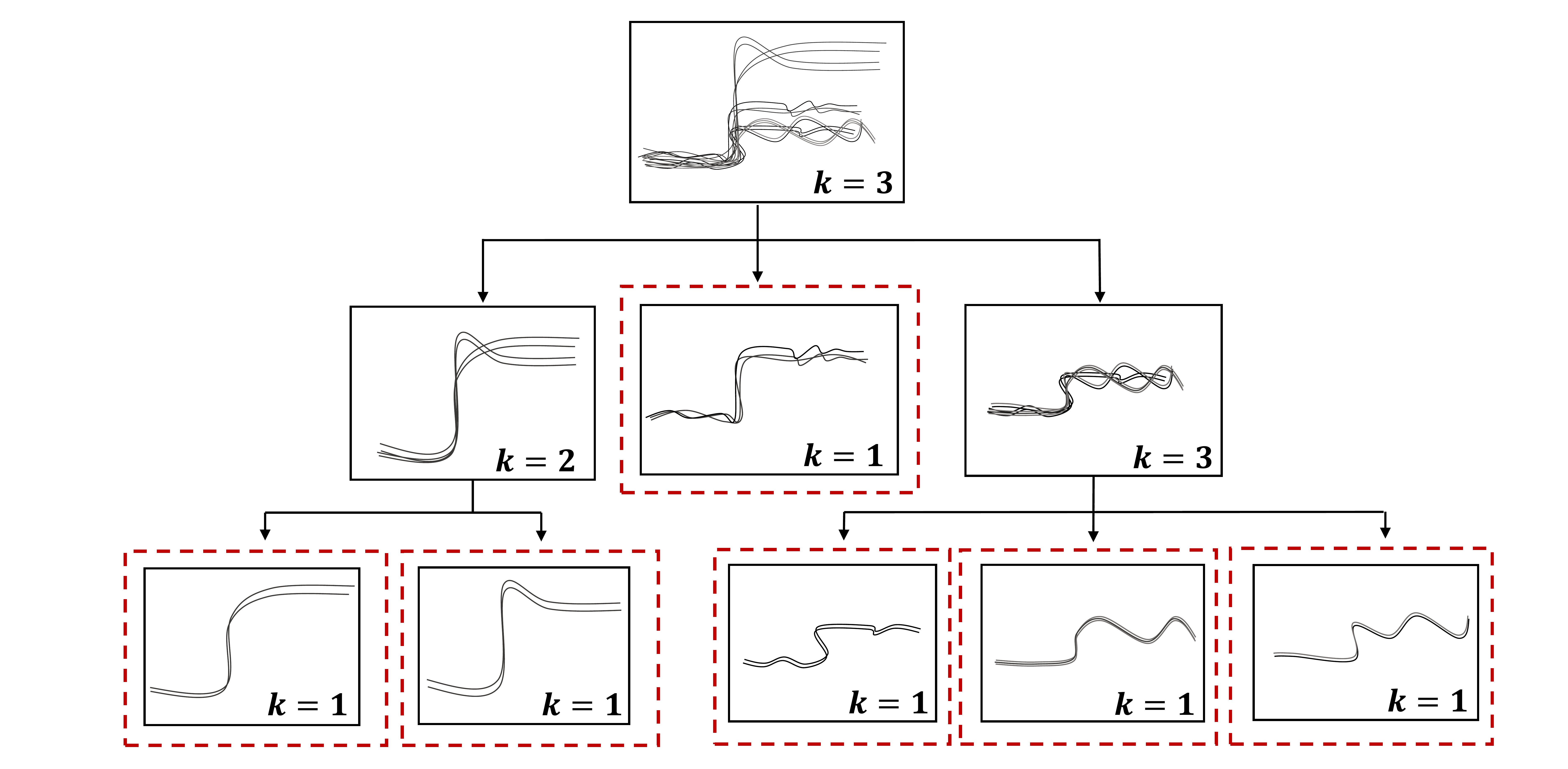}
\caption{\label{fig2}Framework of Iterative Eigengap Search (IES) for discovering patterns in different scales (groups with $k$=1 are accepted as the final clusters)}
\label{fig2}
\end{figure}

\begin{figure}
       \includegraphics[width=0.45\textwidth]{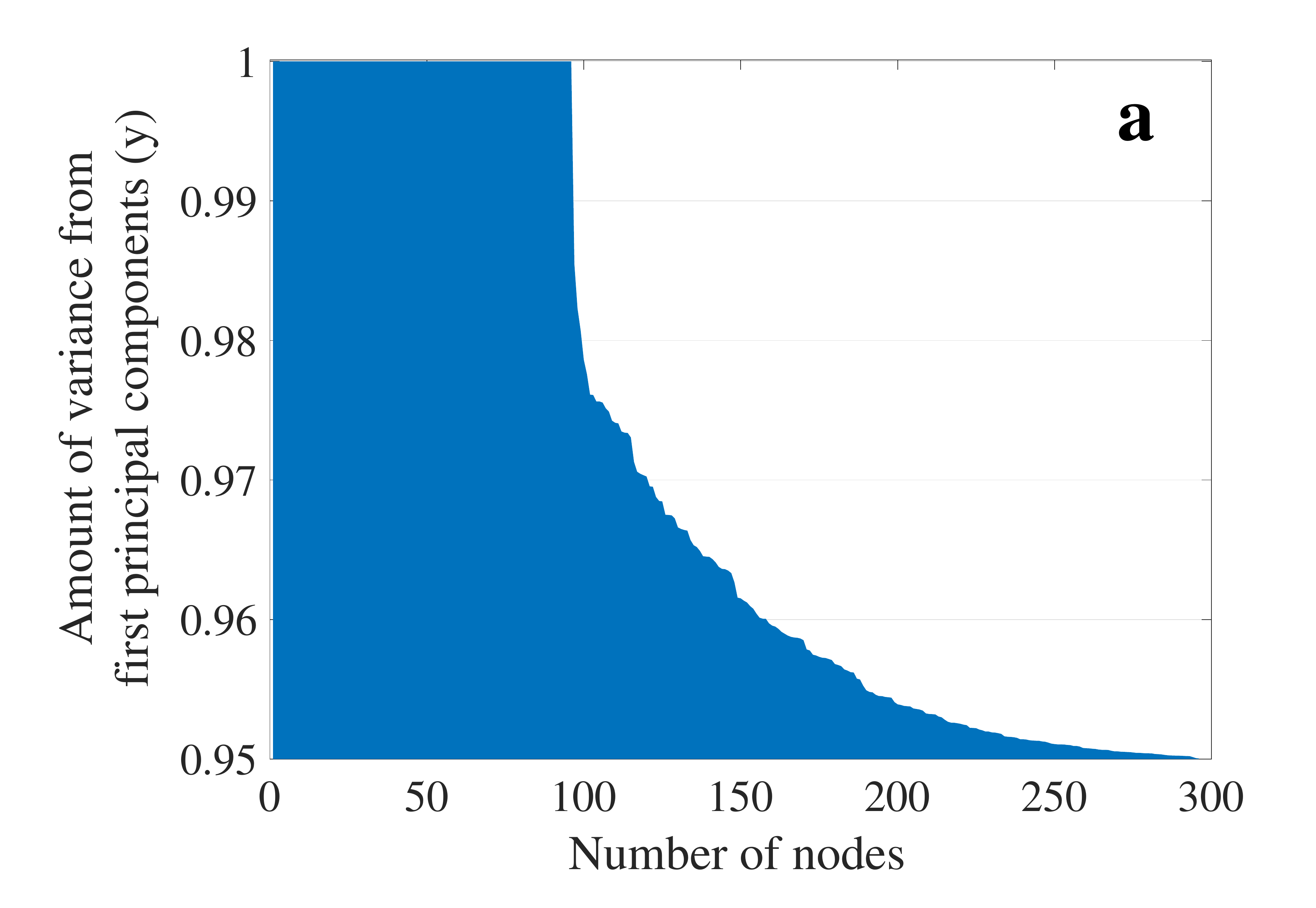}
     \hfill
       \includegraphics[width=0.45\textwidth]{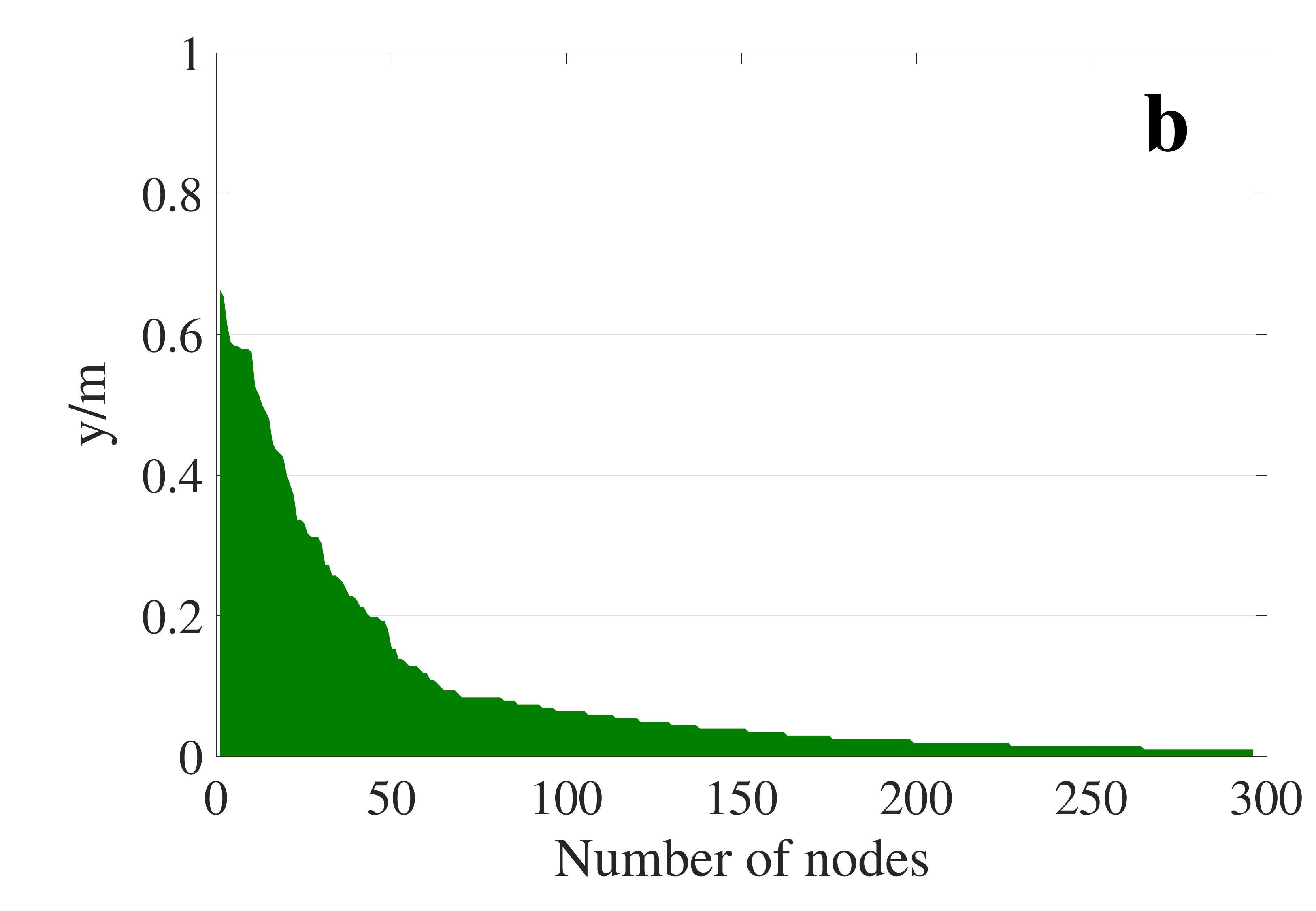}
     \caption{Empirical analysis for the selection of major components in the generated nodes in IES: (a) Amount of variance granted, (b) ratio of selected major components }
     \label{fig3}
   \end{figure}

As described in section \ref{sec311}, we used the inequality in Eq. (\ref{equation8}) to consider the major principal components. Considering the fact that the first major components typically account for the most variance in the data \cite{38}, we select the first major components such that at least 95 percent of variance is granted. As an empirical demonstration, we have considered all the datasets, later described in section \ref{42}, and measured the amount of variance and the percentage of major components for each generated node in the IES, as shown in Fig \ref{fig3}. As can be seen in Fig \ref{fig3}(a), the inequality results in an amount of variance that is close to 1 in our heuristic. On the other hand, as shown in Fig \ref{fig3}(b), in most cases, the number of principal components is limited to a small portion of the features to achieve the objective described in Eq. (\ref{equation8}). Therefore, we used the cut-off threshold of 0.95 to account for almost the whole variance by limiting the number of principal components.

\subsubsection{Iterative Eigengap Search with local scale}
In this alternative of the heuristic, we have adopted the local scaling parameter that identifies the scale by quantifying the distance between nearest neighbors in our search tree framework. As proposed by the original work by Zelnik and Penora \cite{13}, we have utilized 7 nearest neighbors for the  estimation. The selected number of neighbors was suggested in \cite{13} based on comprehensive analysis of both high dimensional and low dimensional data. This approach considers the impact of point-to-point distance in forming the affinity matrix such that multiple scales of data are accounted for. The local scale is used in the Iterative Eigengap Search to identify the structure in the feature space. Given that local scaling has already considered the multi-scale nature of data, the results through one iteration could be considered as the clustering output, which we call eigengap with local scaling (ELS). 
Fig \ref{fig4} presents the pseudo-code for the Iterative Eigengap Search. The tree search is carried out similar to a depth-first search algorithm and therefore a stack data structure that uses the LIFO (last-in-first-out) feature is used to store the data (and the subsequent sub-sections). 

 \begin{figure}
\centering
\includegraphics[width=0.55\textwidth]{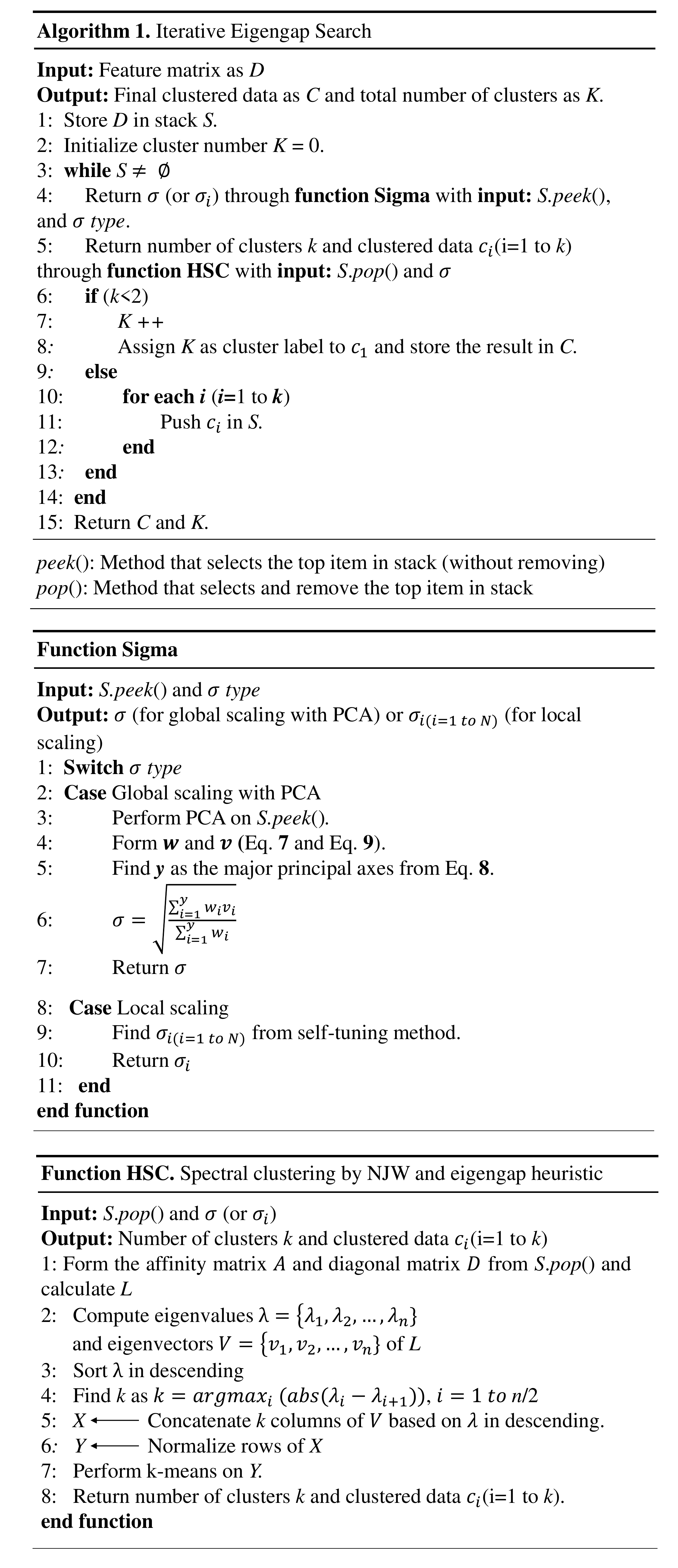}
\caption{\label{fig4}Pseudo-code for Iterative Eigengap Search Heuristic}
\label{fig4}
\end{figure}

In this work, we have adopted NJW, which uses the normalized Laplacian matrix to extract the structure of the data as the standard spectral clustering (SC) for our automated clustering method. In the past recent years, different variations of SC have been proposed that showed improvement over NJW from specific perspectives including improved eigenvector selection \cite{39,40}, alternate affinity matrix generation \cite{41}, and reduced computational cost \cite{42,43}. Nonetheless, we have adopted NJW as a seminal well-established algorithm. Considering the nature of the proposed framework for automated clustering, other variations of spectral clustering could be replaced instead, as long as they employ a graph Laplacian matrix (e.g., \cite{20,44}) that enables the use of eigengap heuristic and the scaling parameter in their similarity estimation.

\section{Algorithm Evaluation} \label{sec4}
We have evaluated the proposed heuristic spectral clustering algorithm on two categories of higher dimensional datasets, namely, power consumption time-series (as described for energy disaggregation) and cell cycle data (representing gene expression). Both types of datasets are of higher dimensions, comprised of different scales, and represent real-world data and therefore exposed to noise. No assumption regarding the number of clusters or scaling parameter is taken into account for the analysis. The evaluation metrics, dataset descriptions, and the results will follow.

\subsection{Evaluation Metrics}
In this study, the algorithm performance has been explored through external validation and was compared with internal validation techniques. Clustering validation is a domain which determines the goodness of clustering output \cite{33}. While external validation relies on the external data such as the class labels, internal validation only searches for the information in the data to check the goodness of partitioning, and can also be employed to find the optimal number of clusters \cite{45}. Both data types used in this study are labeled, which enables us to use external validation. However, we are also checking the performance against commonly used minimization of the sum of squared error in cluster dispersion to contrast the algorithm performance against conventional methods of automated clustering.

In internal validation, different metrics typically consider the compactness (high intra-cluster similarity) and separation (low inter-cluster similarity) to estimate the quality of partitions. These metrics can be used as a measure to find the optimal number of clusters. To measure the dispersion (or tightness) of clusters, the sum of the squared error (SSE) \cite{46,47} can be measured as: 

\begin{equation}
SSE_k = \sum\nolimits_{i} \sum\nolimits_{x \in C_i} \parallel x - \bar{x_i} \parallel ^2  \quad i = 1,2,...,k
\label{equation14}
\end{equation}

where $x$ are the data points in cluster $i$, $\bar{x_i}$ is the centroid of cluster $i$, and $k$ is the total number of clusters.

The SSE is measured for a set of clustering outcome for a range of $k$ values to form an “elbow curve”. The optimal number of cluster is decided based on the rate of dispersion by identifying a threshold for change between subsequent values on the elbow curve. 

For the external validation, as the data is fully labeled, we have adopted precision, recall, and F-measure of the confusion matrix. Based on a majority vote, we assign a dominant label to a cluster and form the confusion matrix. 

\subsection{Dataset Description} \label{42}
\subsubsection{Power consumption datasets}
This category of data is focused on power time-series and the power draw of appliances in a typical building. As appliances change their operational states (e.g., going from off to on), the power draw changes. Clustering has applications in non-intrusive electricity consumption disaggregation, which uses minimal sensing in a building unit coupled with machine learning frameworks. More details on the need and challenges for automated clustering in this field of problems could be found in \cite{48}. In order to shed light on the nature of this data type, Fig \ref{fig5} shows a sample of raw time series data over a 3-hour period with 60Hz resolution. The red circles in this figure illustrate the events, when operational states of appliances were changed.

 \begin{figure}
\centering
\includegraphics[width=0.5\textwidth]{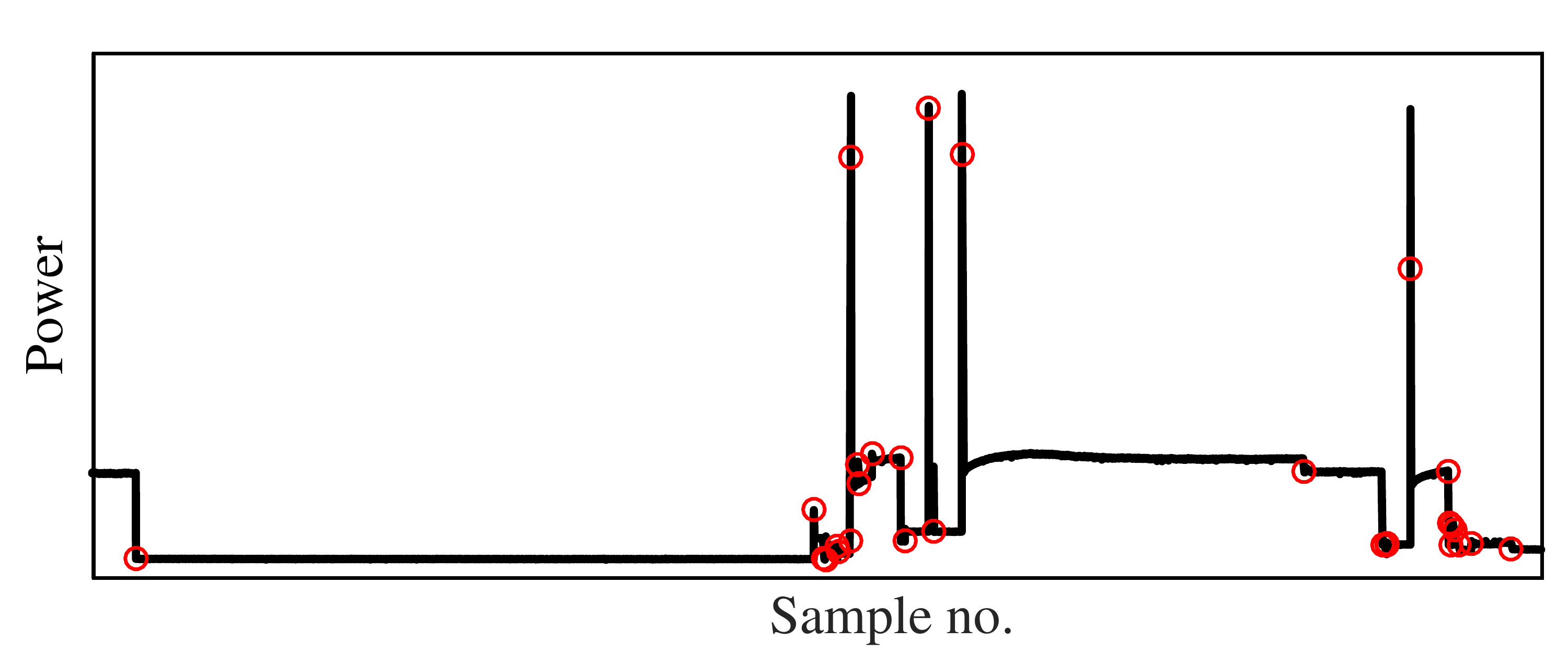}
\caption{\label{fig5}Average of measurement difference for different labels}
\label{fig5}
\end{figure}

 \begin{figure}
\centering
\includegraphics[width=1\textwidth]{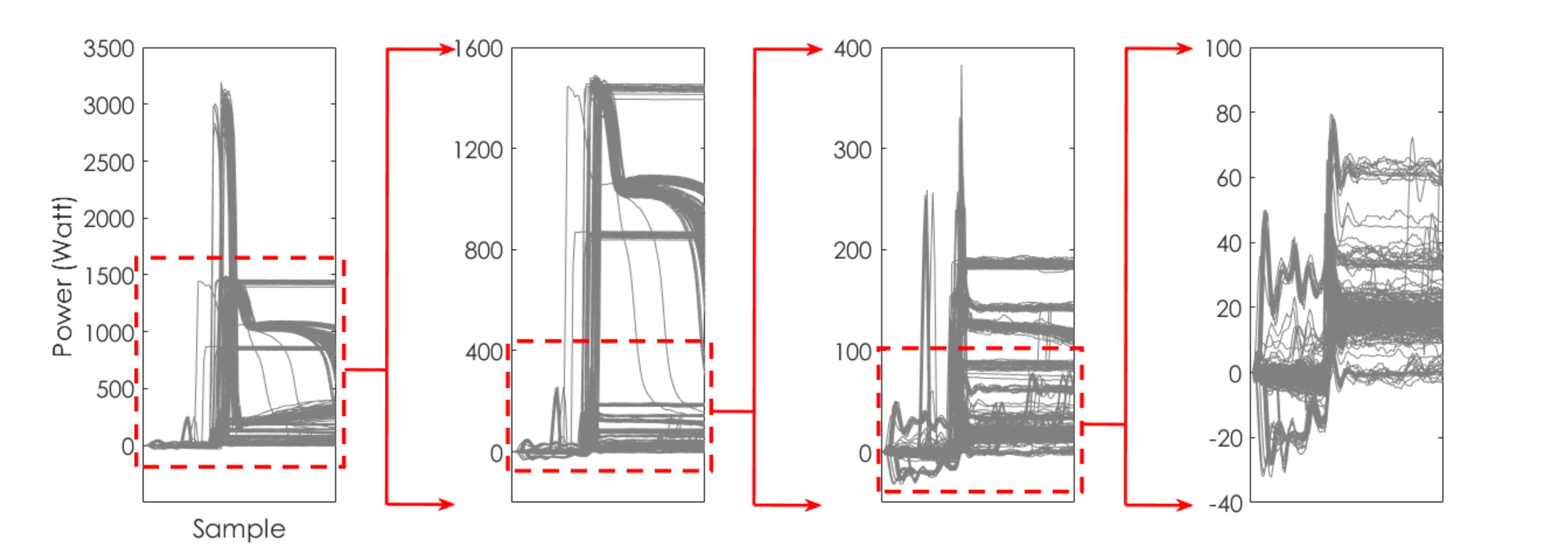}
\caption{\label{fig5}Visualization of a power dataset (time series signal) that shows the effect of multiple scales for the clustering problem; the presence of different clusters are magnified from left to right. In the right frame, 7 groups exist while in the left frame their presence is entirely concealed due to the presence of other patterns with high measurement difference.}
\label{fig6}
\end{figure}

 \begin{figure}
\centering
\includegraphics[width=1.04\textwidth]{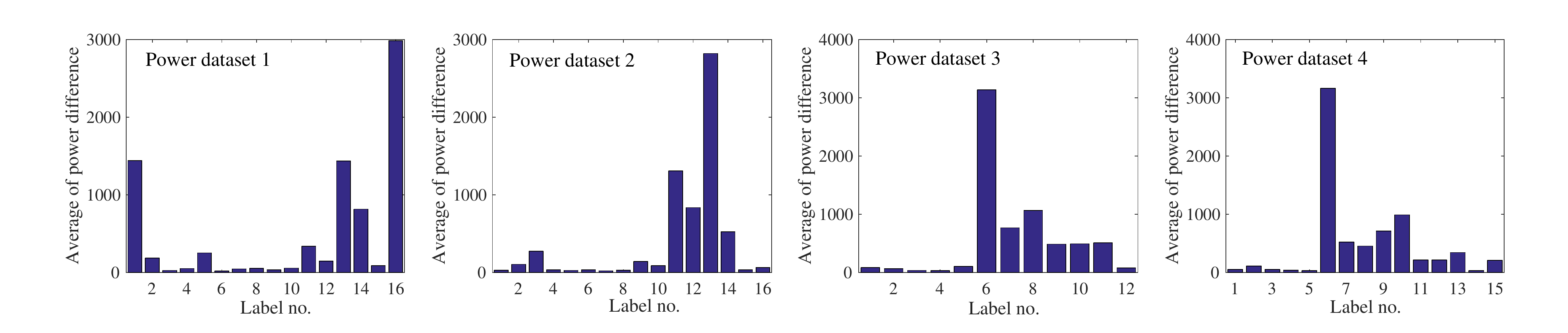}
\caption{\label{fig7}Average of measurement difference for different labels}
\label{fig7}
\end{figure}

Events are detected using the Generalized Likelihood Ratio (GLR) event detection algorithm. Therefore, the dataset contains the noise due to performance of automated feature extraction algorithms as well. The transient in power draw in the vicinity of these events are defined as the appliance signatures and are used as feature vectors, rendering this problem as a feature-based time series clustering according to \cite{49}. In this study, we have used the transient signature (comprised of real and reactive power) for one second after each event and 2/3 of a second before each one. This dataset has been collected and labeled in three occupied apartments over the course of two weeks \cite{50}, in which we used the data from the first apartment for our analysis. The dataset is fully labeled under human supervision with the data from ground truth sensors. The labels represent appliances operational states, and each appliance could have several operational states. These labels have been used for external validation. \newline

Power data has a highly multi-scale nature, which has been visualized in Fig \ref{fig6} for one example dataset. This dataset contains 16 labels (i.e., different classes). In this figure, feature vectors for all instances have been plotted (only real power section of the vectors was presented). Going from left to right, feature vectors in the larger scales were recursively removed and thus the dissimilarities in smaller scales have been revealed. Differences in scales stem from differences in appliances’ power draw. In the smallest scale, the dataset contains 7 clusters that are completely masked when the scaling parameter is not estimated according to that scale. The challenging task of clustering in this problem arises from the fact that automated clustering can simply overlook distinguishing patterns in the small-scale region. In this study, we have used four power datasets, for which in Fig \ref{fig7}, the average of variations for all the events of particular labels were plotted. The wide range of power variations for a multitude of labels in all the datasets demonstrates the multi-scale characteristic of this type of data.

\subsubsection{Gene expression dataset}
While the inspiration for the proposed approach raised from the energy disaggregation problem with highly multi-scale nature, the proposed approach is a generic one and could be used in other domains with similar data types. As the second category of data, we have selected a gene expression data. Clustering is a popular approach for the analysis of gene expression in bioinformatics to study the variation of genes. The dataset has higher dimensions (compared to 2D benchmark problems). We have selected a gene expression data type \cite{51} that displays the fluctuation of the expression level of almost 6000 genes that were collected at 17 time points (resulting in 17 attributes). In \cite{52}, a subset of 384 genes with five phases of cell cycle (five classes) was extracted as a benchmark dataset for clustering, which we used here as one of our examples. 
Table \ref{Table1} provides the description of all the datasets used in this study.

\begin{table}[]
\centering
\begin{tabular}{@{}lccc@{}}
\toprule 
Dataset            & \multicolumn{1}{l}{No. of data points} & \multicolumn{1}{l}{No. of features} & \multicolumn{1}{l}{No. of classes} \\  \midrule 
Power dataset 1    & 756                                    & 202                                 & 16                                 \\
Power dataset 2    & 498                                    & 202                                 & 16                                 \\
Power dataset 3    & 1454                                   & 202                                 & 12                                 \\
Power dataset 4    & 2235                                   & 202                                 & 15                                 \\
Cell cycle dataset & 384                                    & 17                                  & 5                                  \\ \bottomrule
\end{tabular}
\caption{Description of datasets} \label{Table1}
\end{table}

\subsection{Performance Assessment}
In this section, we provide details on qualitative and quantitative assessments. In the former, the effect of the proposed algorithm on the quality of clusters has been visually described. The latter evaluates the performance by using metrics including accuracy, F-measure and computational time. A comparative performance assessment has been also included for the comparison purpose. 

\subsubsection{Qualitative performance assessment}
\begin{figure}
\centering
\includegraphics[width=1\textwidth]{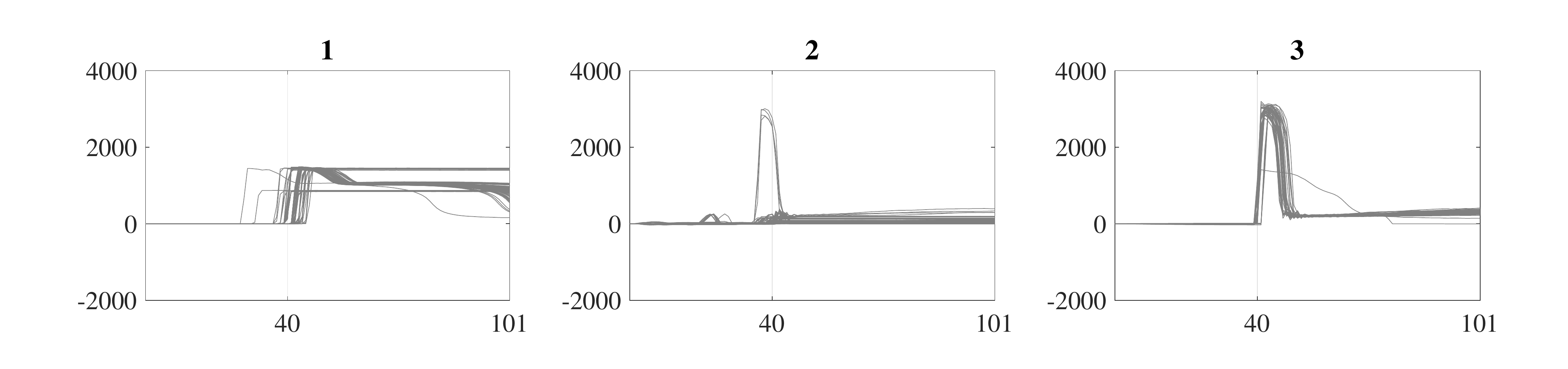}
\caption{\label{fig8}Clustering outcome after first iteration with the coarse-level division on power dataset 1}
\end{figure}

\begin{figure}
\centering
\includegraphics[width=1\textwidth]{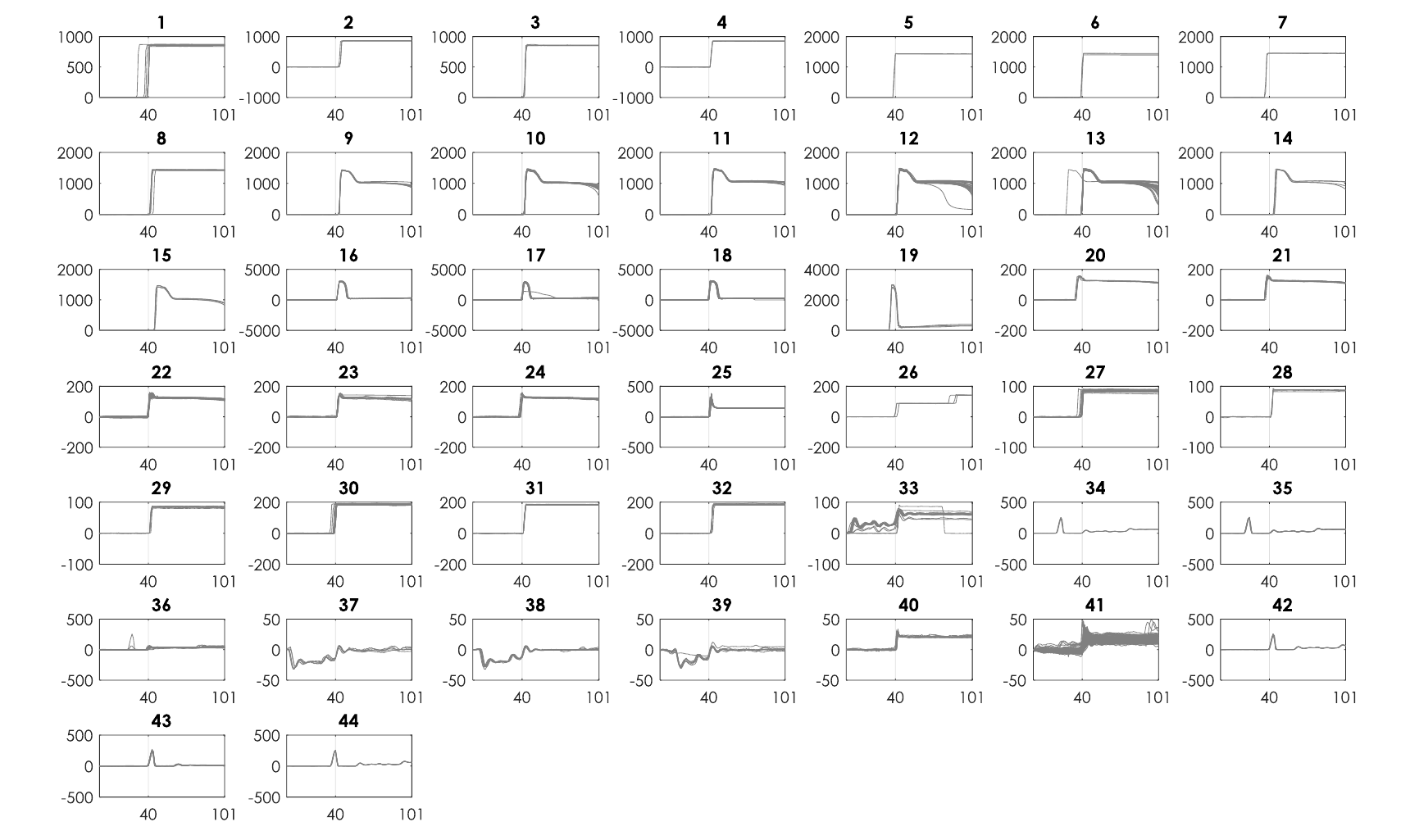}
\caption{\label{fig9}Clustering outcome using Iterative Eigengap Search with \textit{global scaling} on power dataset 1}
\end{figure}

The qualitative assessment is presented for selected datasets that help illustrate (accentuate visual variations) the challenges and performance of the algorithm. Fig \ref{fig8} and Fig \ref{fig9} visualize the clustering output for power dataset 1. Fig \ref{fig8} shows the clustering outcome with Iterative Eigengap Search (IES) with global scaling after the first iteration on the search tree. $K$ is initially estimated as 3, and 3 child nodes are generated. As expected, conventional eigengap is not able to identify the structure of the feature space and resulted in low-quality clusters. Fig \ref{fig9} shows the outcome of the clustering for the iterative search of eigengap, in which 44 clusters were identified at the leaf nodes (where $k$=1). As Fig \ref{fig8} and Fig \ref{fig9} demonstrate, the Iterative Eigengap Search starts with a coarse level separation of clusters in the first iteration and refines the result iteratively to provide high-quality clusters as the outcome.

\begin{figure}
\centering
\includegraphics[width=1\textwidth]{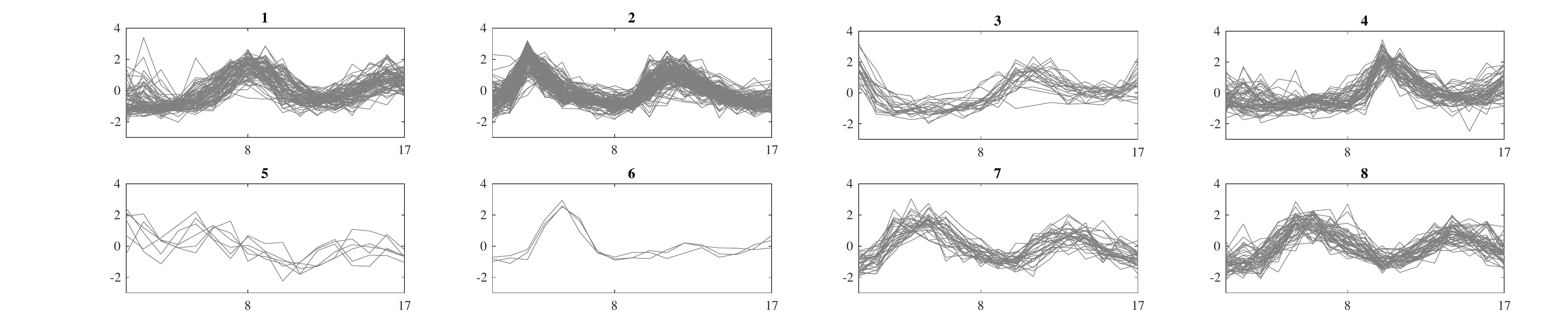}
\caption{\label{fig10}Clustering outcome using Iterative Eigengap Search with global scaling on cell cycle dataset}
\end{figure}

For the cell cycle dataset, Iterative Eigengap Search starts by generating four clusters in the first iteration and generates more clusters in the next level. Due to random initialization of centroids in k-means clustering step (which could result in local optima), the final number of nodes may vary slightly. Fig \ref{fig10} visualizes the cluster results for the cell cycle dataset, where $K$ was found to be 8. 

\begin{figure}[ht]
  \centering
  \begin{subfigure}{\linewidth}
    \centering
    \includegraphics[width=1\linewidth]{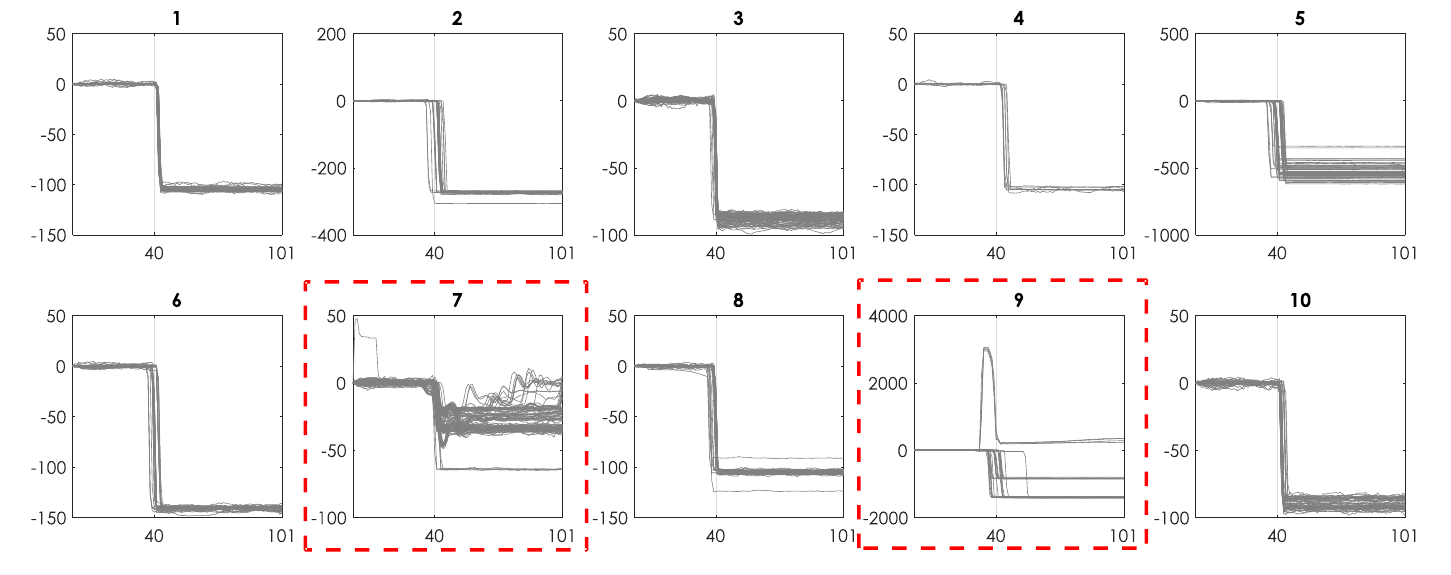}
    \caption{}
  \end{subfigure}

  \begin{subfigure}{\linewidth}
    \centering
    \includegraphics[width=1\linewidth]{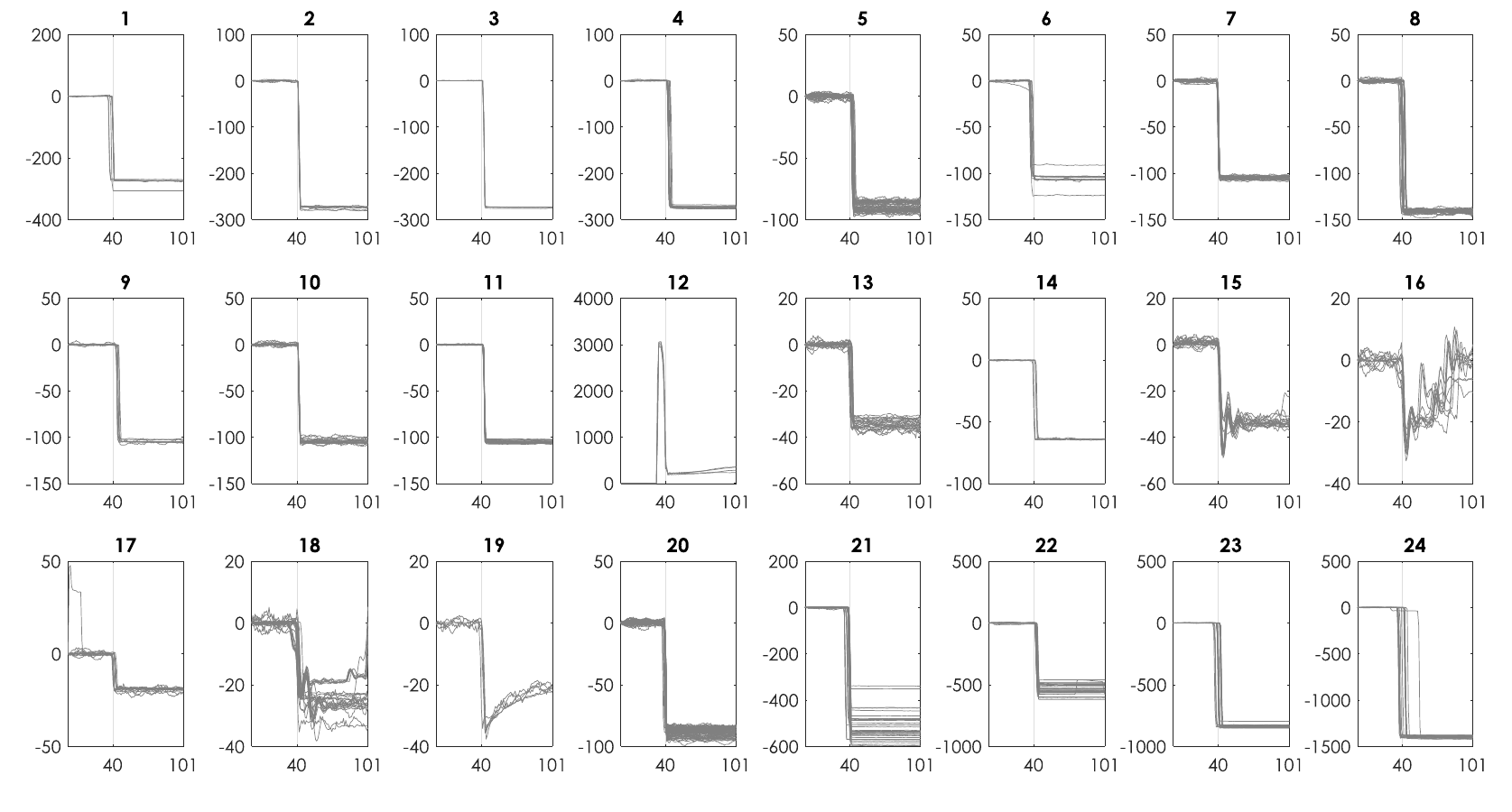}
    \caption{}
  \end{subfigure}  
  \caption{\label{fig11} Clustering outcome using Iterative Eigengap Search with \textit{local scaling} on power dataset 2: (a) cluster outcome after first iteration (clusters with potential for improvement were highlighted) and (b) cluster outcome on the leaf nodes}  
\end{figure}

Fig \ref{fig11} shows the clustering outcome for Iterative Eigengap Search with local scaling on power dataset 2. In Fig \ref{fig11}-a, the clusters in root node were presented. As shown in this figure, a reasonable estimation for $k$ is achieved, but there are few clusters (highlighted with dash lines), which potentially could be improved. Fig \ref{fig11}-b presents how the iterative approach modifies the clusters. Local scaling results in a higher number of clusters in the first iteration with a shallower tree structure. 

The quantitative impact of this difference, both in terms of accuracy and computational time, will follow.

\subsubsection{Quantitative performance assessment}
\begin{figure}
\centering
\includegraphics[width=0.5\textwidth]{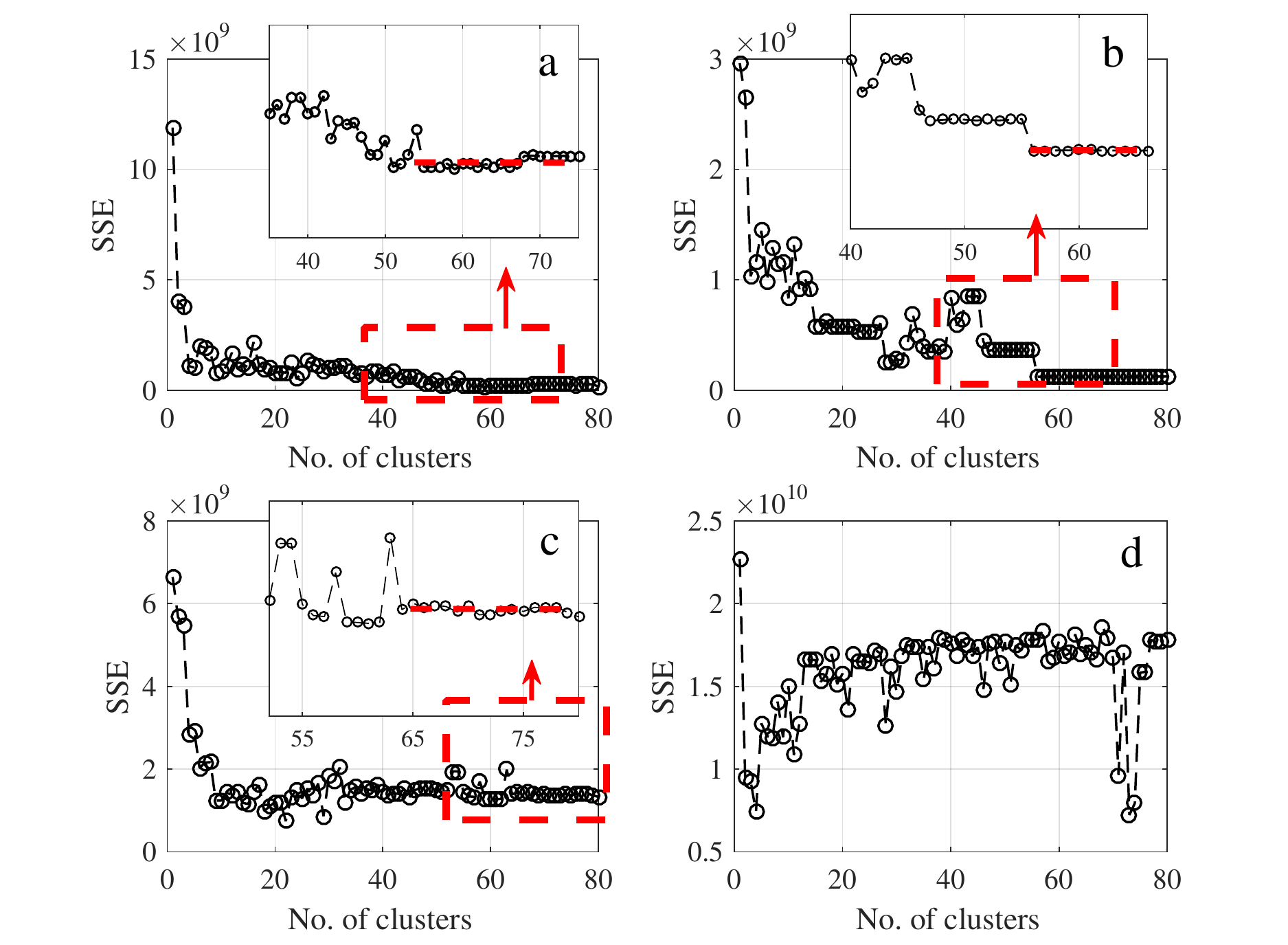}
\caption{\label{fig12}Elbow curves for a) power dataset 1, b) power dataset 2, c) power dataset 3, and d) power dataset 4}
\end{figure}

Given that the labels for data points are known, we carried out external validations to quantify the algorithm performance in comparison to state-of-the-art and conventional internal validation. For internal validation, by considering a range for the number of clusters in ascending order, the SSE was obtained for each dataset using the concept described in Eq. (\ref{equation14}). Fig \ref{fig12} shows the elbow curves for all power datasets.
 In all cases, PCA was used to estimate the scaling factor ($\sigma$). Since spectral clustering algorithm implements K-means in the last step, the results are affected by the random initialization of centroids, and thus the structure of the elbow curve for the subsequent number of $k$’s would be affected. To avoid this bias, we assigned fixed initial seeds values (i.e., the same specific data points for initialization of K-means) for all $k$ values in forming the elbow curve. As shown in Fig \ref{fig12}, the noisy structure of data brings about inconsistencies in descending trend of SSE values as $k$ increases, though the general pattern of flattening for measurement is preserved, except for case (d). For case (d), since the elbow curve within the considered range of $k$ is not presented, estimating $k$ based on this plot has not been considered. Upon forming the elbow curves, we have manually selected the number of clusters by visual evaluation of the rate of decrease in values for the internal validation. 
 
To associate the cluster label with the ground truth, we form a matrix that relates the cluster number (assigned by the algorithm) for each observation to its corresponding ground truth label and call it the association matrix henceforth. As an example, Table \ref{table2} shows the association matrix for power dataset 1. The association matrix is mapped to a confusion matrix for the performance quantification. In forming the confusion matrix, clusters are labeled based on the majority vote. In order to provide insight on labeling clusters with the majority vote, let us consider cluster number 23, which contains 20 feature vectors (i.e., data points) in total. It could be seen that 19 data points are from class 14501 and one instance is from class 18001. Considering the dominance of class 14501, it is assigned as the label for cluster 23.

\begin{table}[]
\centering
\includegraphics[width=0.95\textwidth]{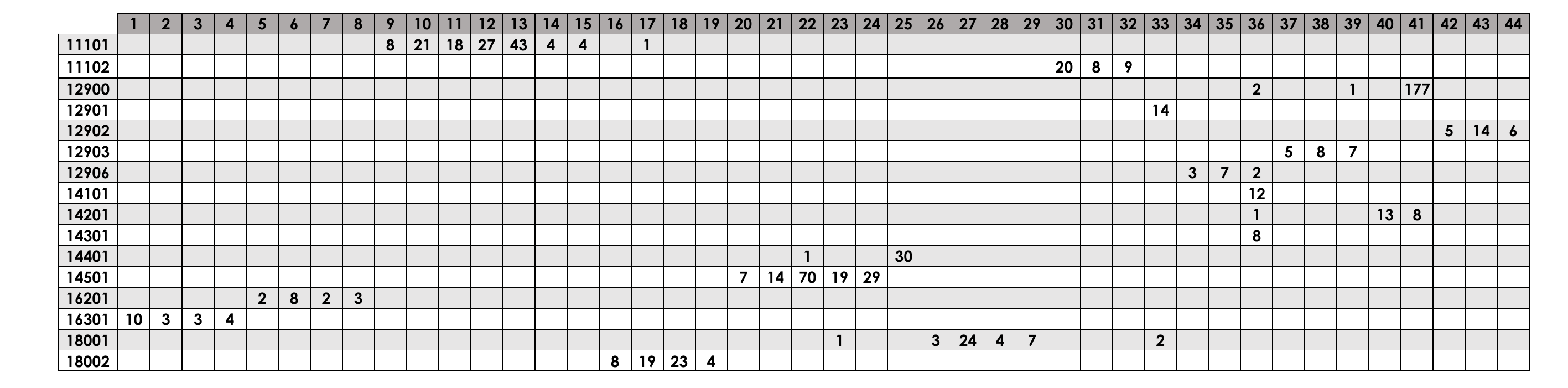}
\caption{\label{table2}Association matrix between generated clusters and ground truth label (power dataset 1)}
\end{table}

\begin{figure}
\centering
\includegraphics[width=0.45\textwidth]{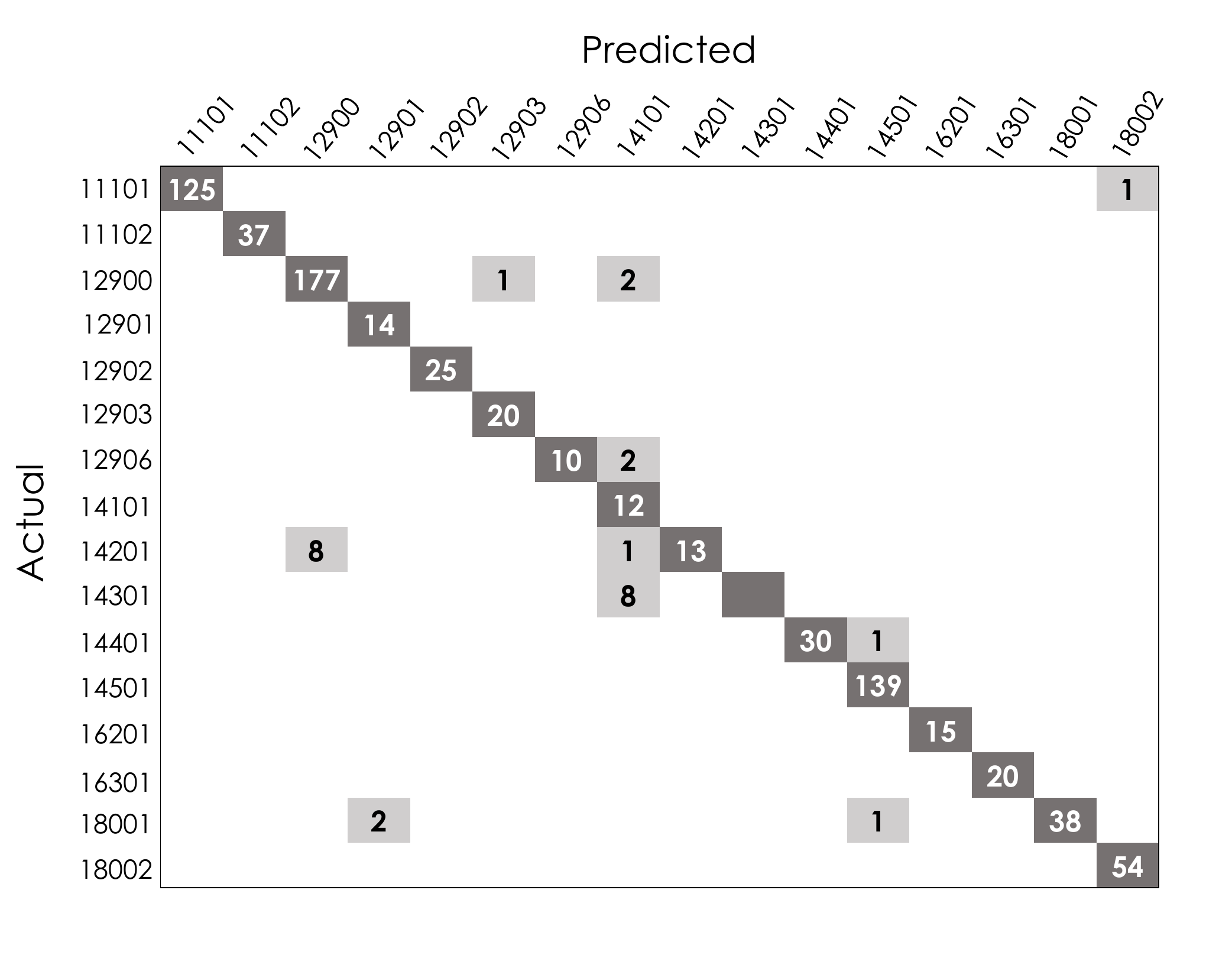}
\caption{\label{fig13}The equivalent confusion matrix mapped from association matrix (power dataset1)}
\end{figure}

Ideally, the number of generated clusters will be the same as the number of ground truth labels. However, as the number of generated clusters exceeds the number of ground truth labels, clusters with the same label will be merged to form the confusion matrix. For example, the contents of column 20 to 24 in Table \ref{table2} are cumulated and the groups are merged since they all represent class 14501. Fig \ref{fig13} shows the output of mapping to form the confusion matrix for power dataset 1.

\begin{figure}
\centering
\includegraphics[width=1\textwidth]{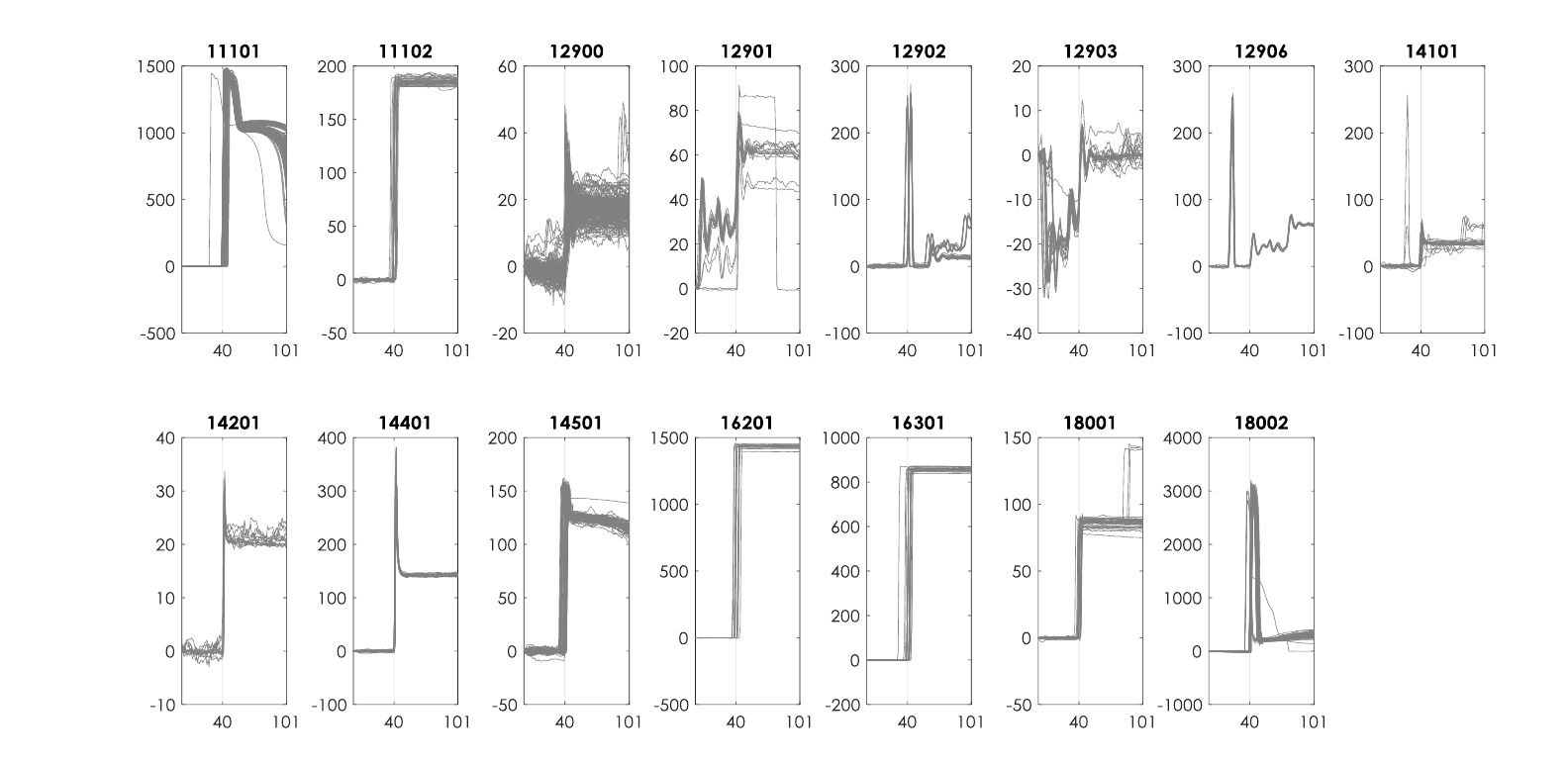}
\caption{\label{fig14}Cluster representation after (manual) merging for power dataset 1. Each label represents an appliance transition state}
\end{figure}

The final clusters after assigning the associated label and (manual) merging of similar clusters is shown in Fig \ref{fig14}. Except for class 14301 that has similar signature representations with 14101, all other classes were retrieved and preserved through the clustering process. It must be noted that the abovementioned process of manual merging was performed only for visualization of clusters in association with the classes in the physical environment. Nonetheless, the quantitative performance assessment of the clustering process was carried out with respect to the direct output of the clustering algorithms through the association matrix (Table \ref{table2}) without any cluster merging.

To demonstrate the effectiveness of the Iterative Eigengap Search (IES) that is the focus of this study, for comparison, we provided the outcome of the following spectral clustering algorithms: (1) similar to our work, the ZP self-tuning technique \cite{13}, and MEG-CD \cite{53} are automated spectral clustering methods, (2) FUSE spectral clustering \cite{54} is specifically focused on multi-scale datasets; (3) and NJW \cite{8} and CPQR-based from \cite{55} are conventional spectral clustering algorithms that call for parameter inputs. In addition, the results for the internal validation, which is commonly used for the validation of clustering methods, as well as the legacy eigengap heuristic have been presented.

Calculated from the association matrix, Table \ref{Table3} presents the performance metrics of different methods on all the datasets. Five-fold cross-validation was used for evaluation, and the average is reported for the accuracy, precision, recall, and F-measure metrics to avoid bias in the quantification of performance metrics. As the values in Table \ref{Table3} indicate, Iterative Eigengap Search with global scaling shows the best performance. Followed by that is the Iterative Eigengap Search with local scaling with fewer number of clusters, which is more compatible with the natural separation of patterns in the feature space. As noted, Iterative Eigengap Search does not require any range for the number of clusters. For the ZP self-tuning \cite{13} and MEG-CD \cite{53}, although considered as automated clustering, they required a range of initial values to optimize over the number of clusters. Therefore, a range of 2 to 80 clusters for power datasets, and 2 to 20 for the cell cycle dataset were considered for ZP self-tuning \cite{13} and internal validation. A range of 100 to 3000 with intervals of 100 for $\sigma$ was considered for MEG-CD \cite{53}. We chose this range based on our empirical observations on the dataset. Since ZP self-tuning \cite{13} underestimates the number of clusters, it does not result in an accurate outcome, specifically for the power datasets. Similarly, legacy eigengap heuristic leads to a smaller number of clusters and consequently low performance due to its incapability of accounting for the multi-scale nature. MEG-CD \cite{53} performed better in terms of estimating the number of clusters but the inaccurate estimation of $\sigma$ (shown by the internal validation) led to a low performance in clustering. For NJW \cite{8}, since both $K$ and $\sigma$ are estimated manually, $K$ is assumed to be equal to the number of classes (input information), and $\sigma$ was selected such that the clusters with smallest distortion are obtained. The results show that while $K$ was manually selected to its true value, the performance in all the cases falls behind the Iterative Eigengap Search. For the recent multi-scale clustering algorithm, FUSE \cite{54}, the IES outperforms as well. Also, CPQR-based from \cite{55} showed to be less accurate compared to the IES in all cases, but the approach has the highest computational efficiency among all the methods. The last column in Table \ref{Table3} shows the total analysis runtime. It must be noted that for the NJW algorithm \cite{8} ZP self-tuning \cite{13}, and MEG-CD \cite{53}, we had to define a range for $\sigma$ and $K$, which consequently affects the reported runtime. Based on the extent of familiarity with the problem, a higher or lower range can be defined, which can significantly change the reported time.

\begin{table}[]
\centering
\resizebox{\textwidth}{!}{%
\begin{tabular}{@{}llccccccc@{}}

\toprule
Method                                                                                                   & Dataset         & \multicolumn{1}{l}{$\sigma$ selection}                                          & $K$   & Accuracy$^{\text{a}}$ & Precision$^{\text{a}}$ & Recall$^{\text{a}}$ & F-measure$^{\text{a}}$ & Runtime(s)   \\ \midrule
\multirow{5}{*}{\begin{tabular}[c]{@{}l@{}}IES with Global\\ Scale (leaf node \\ clusters)\end{tabular}} & Power data1     & \multirow{5}{*}{PCA}                                                         & 44  & \textbf{0.97}     & \textbf{0.97}      & \textbf{0.97}   & \textbf{0.96}      & 15           \\
                                                                                                         & Power data2     &                                                                              & 40  & \textbf{0.89}     & \textbf{0.85}      & \textbf{0.89}   & \textbf{0.86}      & 9            \\
                                                                                                         & Power data3     &                                                                              & 38  & \textbf{0.97}     & \textbf{0.96}      & \textbf{0.97}   & \textbf{0.96}      & 24           \\
                                                                                                         & Power data4     &                                                                              & 60  & \textbf{0.96}     & \textbf{0.95}      & \textbf{0.96}   & \textbf{0.95}      & 78           \\
                                                                                                         & Cell Cycle data &                                                                              & 8   & 0.70     & 0.71      & 0.70   & 0.68      & 0.4          \\ \midrule
\multirow{5}{*}{\begin{tabular}[c]{@{}l@{}}One step IES with \\ Local Scale (ESL)\end{tabular}}          & Power data1     & \multirow{5}{*}{\begin{tabular}[c]{@{}c@{}}Local \\ Scaling\end{tabular}}    & 27  & 0.88     & 0.84      & 0.88   & 0.85      & 6            \\
                                                                                                         & Power data2     &                                                                              & 10  & 0.84     & 0.78      & 0.84   & 0.80      & 4            \\
                                                                                                         & Power data3     &                                                                              & 25  & 0.93     & 0.89      & 0.93   & 0.90      & 8            \\
                                                                                                         & Power data4     &                                                                              & 17  & 0.93     & 0.89      & 0.93   & 0.91      & 13           \\
                                                                                                         & Cell Cycle data$^{\text{b}}$ &                                                                              & 4   & 0.67     & 0.58      & 0.67   & 0.62      & 1.6          \\ \midrule
\multirow{5}{*}{\begin{tabular}[c]{@{}l@{}}IES with Local \\ Scale (leaf node \\ clusters)\end{tabular}} & Power data1     & \multirow{5}{*}{\begin{tabular}[c]{@{}c@{}}Local \\ Scaling\end{tabular}}    & 30  & 0.90     & 0.87      & 0.90   & 0.88      & 55           \\
                                                                                                         & Power data2     &                                                                              & 24  & 0.89     & 0.84      & 0.89   & 0.85      & 43           \\
                                                                                                         & Power data3     &                                                                              & 28  & 0.93     & 0.89      & 0.93   & 0.91      & 50           \\
                                                                                                         & Power data4     &                                                                              & 20  & 0.94     & 0.90      & 0.94   & 0.92      & 47           \\
                                                                                                         & Cell Cycle data &                                                                              & 4   & 0.67     & 0.58      & 0.67   & 0.62      & 6.9          \\ \midrule
\multirow{5}{*}{\begin{tabular}[c]{@{}l@{}}Internal \\ validation\end{tabular}}                          & Power data1     & \multirow{5}{*}{PCA}                                                         & 55  & 0.93     & 0.89      & 0.92   & 0.91      & 89$^{\text{c}}$           \\
                                                                                                         & Power data2     &                                                                              & 56  & 0.87     & 0.83      & 0.87   & 0.85      & 40           \\
                                                                                                         & Power data3     &                                                                              & 45  & 0.94     & 0.94      & 0.94   & 0.93      & 201          \\
                                                                                                         & Power data4     &                                                                              & N/A$^{\text{d}}$ & N/A      & N/A       & N/A    & N/A       & 1201         \\
                                                                                                         & Cell Cycle data &                                                                              & 14  & \textbf{0.73}     & \textbf{0.73}      & \textbf{0.73}   & \textbf{0.72}      & 4.5          \\ \midrule
\multirow{5}{*}{\begin{tabular}[c]{@{}l@{}}Legacy\\ Eigengap\end{tabular}}                               & Power data1     & \multirow{5}{*}{PCA}                                                         & 3   & 0.44     & 0.25      & 0.44   & 0.30      & 1            \\
                                                                                                         & Power data2     &                                                                              & 3   & 0.37     & 0.17      & 0.3    & 0.22      & \textless{}1 \\
                                                                                                         & Power data3     &                                                                              & 6   & 0.60     & 0.36      & 0.59   & 0.44      & 5            \\
                                                                                                         & Power data4     &                                                                              & 5   & 0.51     & 0.26      & 0.51   & 0.35      & 15           \\
                                                                                                         & Cell Cycle data &                                                                              & 4   & 0.66     & 0.58      & 0.66   & 0.62      & 1            \\ \midrule
\multirow{5}{*}{NJW \cite{8}}                                                                                     & Power data1     & \multirow{5}{*}{\begin{tabular}[c]{@{}c@{}}Least \\ Distortion\end{tabular}} & 16$^{\text{e}}$  & 0.82     & 0.73      & 0.82   & 0.76      & 30$^{\text{c}}$           \\
                                                                                                         & Power data2     &                                                                              & 16  & 0.66     & 0.61      & 0.66   & 0.61      & 20           \\
                                                                                                         & Power data3     &                                                                              & 12  & 0.89     & 0.82      & 0.89   & 0.85      & 115          \\
                                                                                                         & Power data4     &                                                                              & 15  & 0.54     & 0.30      & 0.54   & 0.38      & 520          \\
                                                                                                         & Cell Cycle data &                                                                              & 5   & 0.63     & 0.55      & 0.63   & 0.58      & 6            \\ \midrule
\multirow{5}{*}{MEG-CD \cite{53}}                                                                                  & Power data1     & \multirow{5}{*}{\begin{tabular}[c]{@{}c@{}}Local \\ Scaling\end{tabular}}    & 12  & 0.47     & 0.29      & 0.47   & 0.34      & 7$^{\text{c}}$            \\
                                                                                                         & Power data2     &                                                                              & 6   & 0.32     & 0.13      & 0.32   & 0.17      & 4            \\
                                                                                                         & Power data3     &                                                                              & 7   & 0.58     & 0.35      & 0.58   & 0.43      & 34           \\
                                                                                                         & Power data4     &                                                                              & 3   & 0.51     & 0.26      & 0.51   & 0.35      & 111          \\
                                                                                                         & Cell Cycle data &                                                                              & 2   & 0.48     & 0.30      & 0.48   & 0.35      & 2            \\ \midrule
\multirow{5}{*}{FUSE \cite{54}}                                                                                    & Power data1     & \multirow{5}{*}{\begin{tabular}[c]{@{}c@{}}Local \\ Scaling\end{tabular}}    & 16$^{\text{e}}$  & 0.76     & 0.66      & 0.76   & 0.69      & 27           \\
                                                                                                         & Power data2     &                                                                              & 16  & 0.68     & 0.53      & 0.68   & 0.58      & 19           \\
                                                                                                         & Power data3     &                                                                              & 12  & 0.86     & 0.78      & 0.86   & 0.81      & 17           \\
                                                                                                         & Power data4     &                                                                              & 15  & 0.94     & 0.90      & 0.94   & 0.92      & 50           \\
                                                                                                         & Cell Cycle data &                                                                              & 5   & 0.42     & 0.24      & 0.42   & 0.30      & 1.3          \\ \midrule
\multirow{5}{*}{CPQR-based from \cite{55}}                                                                                    & Power data1     & \multirow{5}{*}{\begin{tabular}[c]{@{}c@{}}Local \\ Scaling\end{tabular}}    & 16$^{\text{e}}$  & 0.80     & 0.76      & 0.80   & 0.77      & 2            \\
                                                                                                         & Power data2     &                                                                              & 16  & 0.83     & 0.76      & 0.83   & 0.80      & 1            \\
                                                                                                         & Power data3     &                                                                              & 12  & 0.86     & 0.75      & 0.86   & 0.80      & 2            \\
                                                                                                         & Power data4     &                                                                              & 15  & 0.91     & 0.84      & 0.91   & 0.87      & 4            \\
                                                                                                         & Cell Cycle data &                                                                              & 5   & 0.44     & 0.26      & 0.44   & 0.32      & 1.3          \\ \midrule
\multirow{5}{*}{\begin{tabular}[c]{@{}l@{}}Self-tuning \\ ZP \cite{13}\end{tabular}}                                                                       & Power data1     & \multirow{5}{*}{\begin{tabular}[c]{@{}c@{}}Local\\ Scaling\end{tabular}}     & 3   & 0.47     & 0.23      & 0.45   & 0.30      & 445$^{\text{c}}$          \\
                                                                                                         & Power data2     &                                                                              & 2   & 0.40     & 0.16      & 0.40   & 0.23      & 225          \\
                                                                                                         & Power data3     &                                                                              & 4   & 0.81     & 0.67      & 0.81   & 0.73      & 584          \\
                                                                                                         & Power data4     &                                                                              & 4   & 0.92     & 0.85      & 0.92   & 0.88      & 850          \\
                                                                                                         & Cell Cycle data &                                                                              & 4   & 0.63     & 0.53      & 0.63   & 0.57      & 3    \\ \bottomrule     
\end{tabular}}
\begin{tablenotes}
      \small
      \item ${}^{a}$ These values are calculated based on the weighted average of each label.
            \item ${}^{b}$ For this dataset, only one iteration is observed by using a local scaling parameter.
            \item ${}^{c}$ Results of these columns with these methods are affected by the range of the considered parameters. 
            \item ${}^{d}$ Not applicable since the elbow curve structure is not formed in the identified range.
            \item ${}^{e}$ $K$ for this approach is manually set to the number of classes since $K$ needs to be known in advance.
    \end{tablenotes}
\caption{Performance quantification of different methods (Performance metrics are averaged over 5-fold cross-validation).} \label{Table3}
\end{table}

\begin{figure}[ht]
  \centering
  \begin{subfigure}{\linewidth}
    \centering
    \includegraphics[width=1\linewidth]{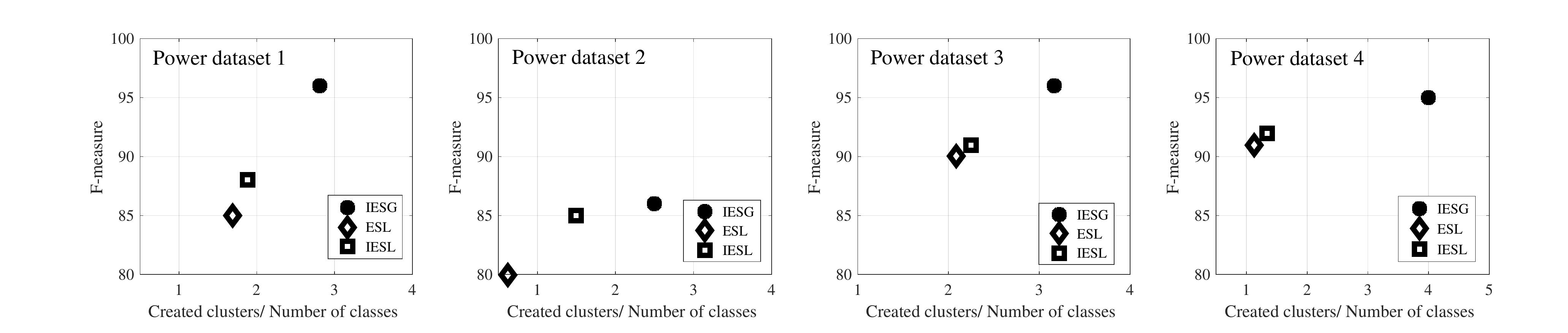}
    \caption{}
  \end{subfigure}

  \begin{subfigure}{\linewidth}
    \centering
    \includegraphics[width=1\linewidth]{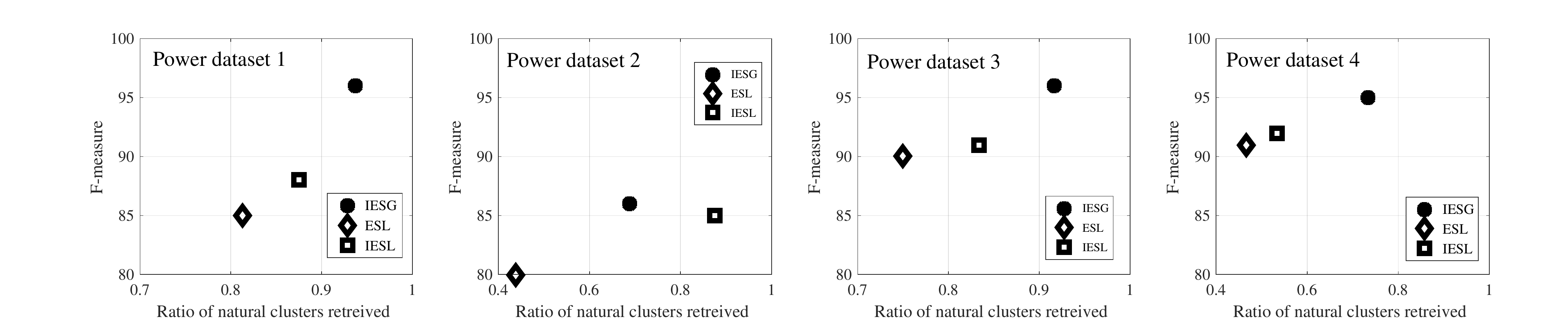}
    \caption{}
  \end{subfigure}  
  \caption{\label{fig15} Variation of cluster quality indicators versus F-measure for (a) generated number of clusters, (b) ability for retrieving natural patterns. For the legend of this plot, IESG denotes IES with global scale (leaf node clusters); ESL denotes one step IES with local scale; IESL denotes IES with local scale (leaf node clusters)}  
\end{figure}

To provide a more accurate context for comparing the clustering outcome, two important factors of clustering quality were taken into account: (1) the ratio of the generated clusters to the number of class labels and (2) the ratio of the class labels retrieved after clustering. The former indicator shows how close the number of clusters is to the number of ground truth labels. Ideally, this value is equal to 1 when all the ground truth observations of each class are contained in one distinct cluster. The latter indicator denotes the percentage of class labels that possess a separate cluster after forming the confusion matrix. A value of 1 indicates the ideal case. However, the similarity between different classes and their significant unbalanced distribution can reduce this value (e.g., in a case where instances of a very small class are put in a cluster that also contains a ratio of a very large class, the majority vote selects the larger class). Fig \ref{fig15} presents the variation of these indicators versus F-measure for power datasets only. Each point in Fig \ref{fig15} represents one of the power datasets. As shown in Fig \ref{fig15}-a, IES with local scaling maintains a better balance between the 1st indicator and the performance. On the other hand, IES with global scaling results in better performance for all the cases at the cost of generating a larger number of clusters. Regardless of the number of clusters, the application of PCA for estimation of the global scaling factor results in improved performance. Considering the 2nd indicator, as shown in Fig \ref{fig15}-b, IES with global scaling outperforms in recalling class labels with high F-measure values (three out of four cases), which could be interpreted as the ability to retrieve natural patterns. 

\begin{figure}
\centering
\includegraphics[width=0.55\textwidth]{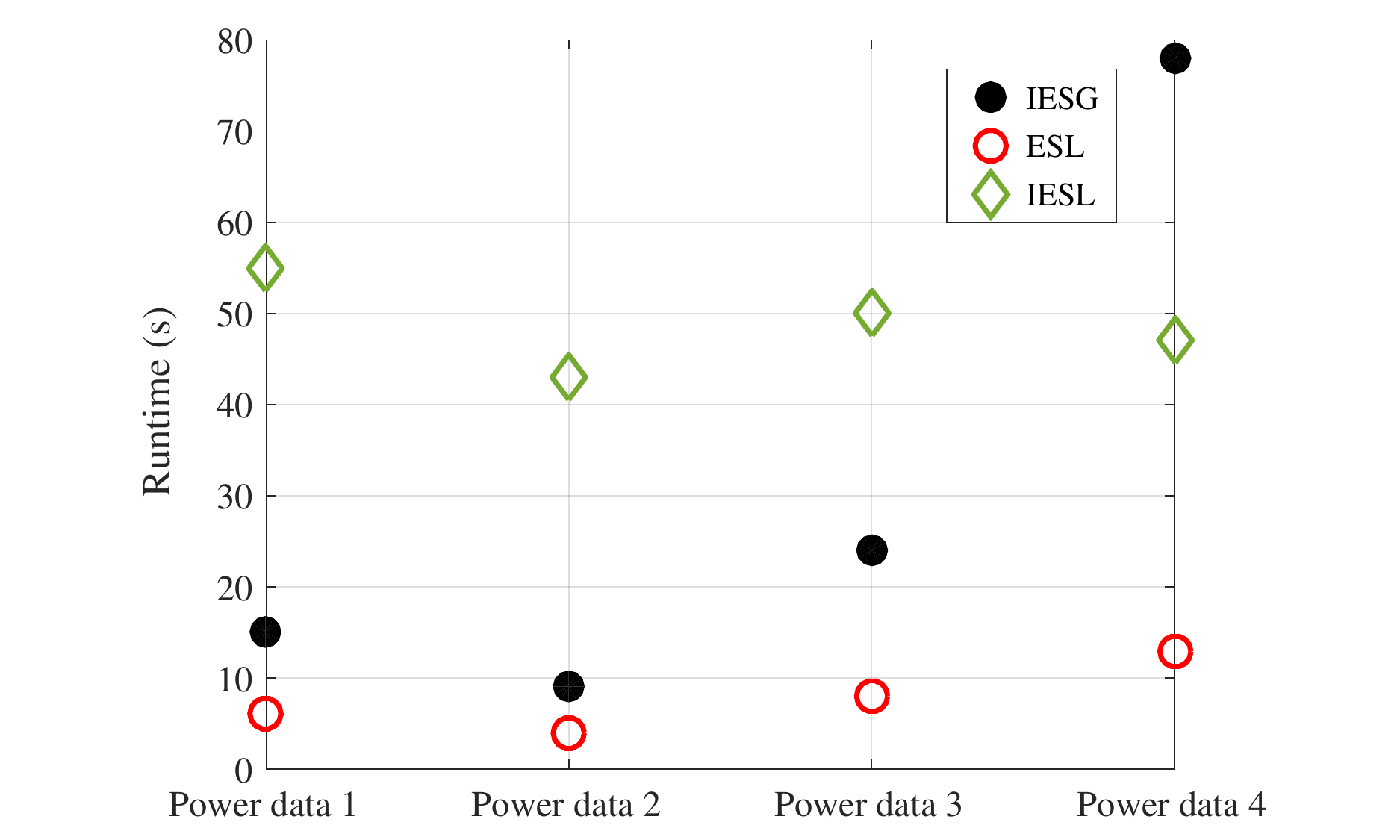}
\caption{\label{fig16} Comparison of analysis runtime}
\end{figure}

In order to provide insight on the computational cost of these techniques, Fig \ref{fig16} presents the run-time for different methods. All the analyses were carried out through MATLAB implementation. As shown in this figure, eigengap search (only one iteration) with local scaling is the most computationally effective approach but it sacrifices the efficacy of results (as discussed from Fig \ref{fig15}(a) and (b) and Table \ref{Table3}). As expected, IES with local scaling generally takes more time compared to IES with global scaling. The result of the comparable self-tuning approach \cite{13} is excluded here to avoid bias since the publically available code was partially implemented with C++, which is known to be more efficient compared to MATLAB. However, since it considers a range of number for clustering as the post-processing step, the results are more computationally expensive unless a narrow range based on the knowledge of the domain is selected. In general, spectral clustering methods are very applicable in different domains. However, the computational cost can get high given the inherent $\mathcal{O}(n^3)$ complexity for eigenvector decomposition. To further improve the computational efficiency, different approximation methods could be adopted for large-scale datasets (e.g., \cite{26,42,43}). 

\begin{figure}
\centering
\includegraphics[width=0.55\textwidth]{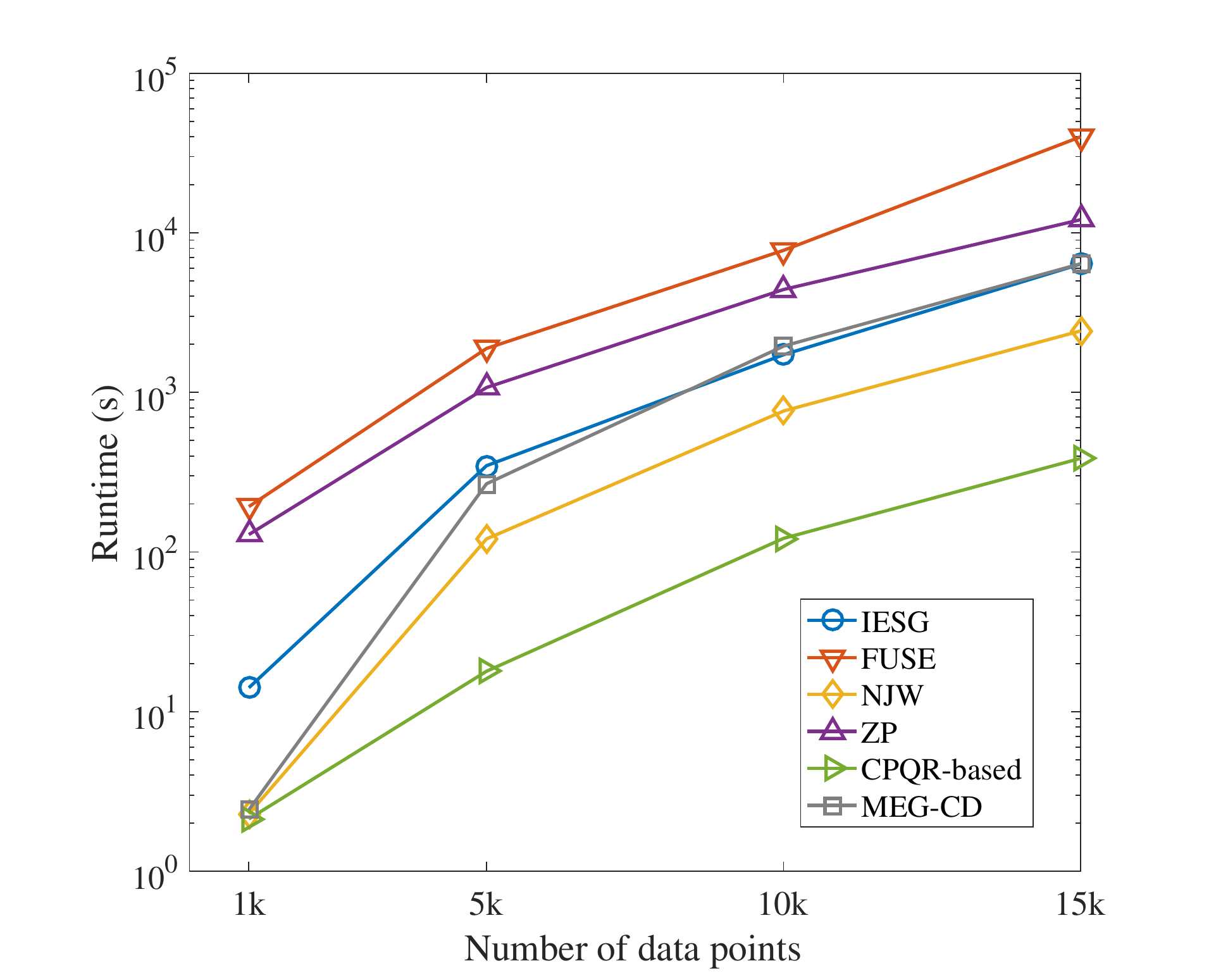}
\caption{\label{fig17} Runtime across a range of datasets with varying number of observations}
\end{figure}

To reflect on the computational efficiency and evaluate the impact of the dataset size on the runtime, we performed an experiment on larger datasets as well. To this end, we merged all power datasets and randomly sampled data points to construct datasets with the desired sizes. In cases that the size of the desired dataset was larger than the mentioned merged dataset, we created synthetic data observations by adding white noise to the original data points for each class. Increasing the number of data points in each class was proportional to the original size of each class. Different datasets with the sizes of 1K, 5K, 10K, and 15K were created for this experiment. All the analyses in this study were carried out on an Intel Xeon CPU 3.5 GHz with 16 GB ram through MATLAB implementation. We have provided the results in Fig \ref{fig17} for different algorithms. For FUSE \cite{54},  NJW \cite{8}, and the CPQR-based from \cite{55}, the number of clusters was set to the number of classes since these methods do not automatically estimate $K$. For self-tuning ZP \cite{13}, as it requires an optimization process over a range of $K$, we used a range of $\pm5$ numbers with respect to the actual number of classes as the estimation  (i.e., the range includes 10 values in addition to the selected one by the algorithm). Similarly, for MEG-CD \cite{53}, an optimization over a range of $\sigma$ is required in which we considered 5 values higher and 5 values lower compared to the optimal selected value by the algorithm. As shown, the runtime of IESG was lower compared to the FUSE \cite{54} and ZP \cite{13}, and almost similar to MEG-CD \cite{53} for larger datasets. The higher runtime of IESG compared to the NJW \cite{8} is intuitive and due to the fact that NJW was used as the standard SC in our search tree framework for automated clustering at each node and was run iteratively on different segments of the data. The analysis involved for each node included the $\sigma$ estimation with PCA analysis and eigengap determination. Also, CPQR-based from \cite{55}, which calls for the number of clusters as an input, had the highest efficiency in the runtime compared to the others. Comparing with the automated or multi-scale methods that call for an optimization process (i.e., ZP, FUSE, and MEG-CD), IESG shows to be practical and scalable in terms of computational efficiency on a larger amount of data.

\section{Conclusion}
We have proposed and evaluated an Iterative Eigengap Search (IES) heuristic for automated spectral clustering of multi-scale and higher dimensional feature spaces. The proposed heuristic does not require \textit{a priori} assumption for the number of clusters ($K$) or scaling parameter of affinity measures ($\sigma$) including a range of values for the number of clusters. The algorithm iteratively searches for eigengaps at different scales of the feature space along a tree structure to partition and refine generated clusters with eigengap heuristic. The scaling parameter is estimated through data-driven methods using (1) a PCA-based global scaling factor or (2) using a local-scaling factor that quantifies local scales by measuring the distance of each data point with its nearest neighbors. The scaling parameters are updated at each node of the tree to reveal the dissimilarities in the local structure of a feature space. We have evaluated the performance of the proposed heuristic on several real-world datasets with multiple classes. The datasets are of higher dimensions with multi-scale and heterogeneous nature. The performance of the IES has been compared against several well-known fundamental spectral clustering methods and an internal validation approach that seeks to minimize the dispersion of clustering outcome. The performance assessments showed that the IES heuristic outperforms comparable approaches in terms of accuracy (an average of 90\% for most of the evaluated cases) and capability of finding (recovering) natural partitions in a feature space.

\bibliographystyle{unsrtnat}
\bibliography{template.bib}

\end{document}